\documentclass[pdflatex,sn-mathphys-num]{sn-jnl}
\usepackage{graphicx}
\usepackage{xcolor}
\usepackage{amsmath}
\usepackage{makecell}
\usepackage{subcaption}
\usepackage{multirow}
\usepackage{booktabs}
\usepackage{amssymb}
\usepackage{array}
\usepackage{hyperref}
\usepackage{microtype}
\usepackage{parskip}
\usepackage{hhline}
\usepackage{tikz}
\usepackage[normalem]{ulem}

\begin{document}
%%%%%%%%%%%%%%%%%%%%%%%%%%%%%%%%%%%%%%%%%%%%%%%%%%%%%%%%%%%
%%%%%%%%%%%%%%%%%%%%    Title    %%%%%%%%%%%%%%%%%%%%%%%%%%
%%%%%%%%%%%%%%%%%%%%%%%%%%%%%%%%%%%%%%%%%%%%%%%%%%%%%%%%%%%

\title[]{Hardware-Aware Tensor Networks for Real-Time Quantum-Inspired Anomaly Detection at Particle Colliders}

\author*[1]{\fnm{Sagar} \sur{Addepalli}}\email{sagar@slac.stanford.edu}
\author[1]{\fnm{Prajita} \sur{Bhattarai}}
\author[1]{\fnm{Abhilasha} \sur{Dave}}
\author[1]{\fnm{Julia} \sur{Gonski}}


\affil[1]{\orgdiv{SLAC National Accelerator Laboratory}, \orgaddress{\street{2575 Sand Hill Rd}, \city{Menlo Park}, \postcode{94025}, \state{CA}, \country{USA}}}

\vspace{10pt}



\maketitle
%

%%%%%%%%%%%%%%%%%%%%%%%%%%%%%%%%%%%%%%%%%%%%%%%%%%%%%%%%%%%
%%%%%%%%%%%%%%%%%%%%    Abstract    %%%%%%%%%%%%%%%%%%%%%%%
%%%%%%%%%%%%%%%%%%%%%%%%%%%%%%%%%%%%%%%%%%%%%%%%%%%%%%%%%%%

\begin{abstract}

Quantum machine learning offers the ability to capture complex correlations in high-dimensional feature spaces, crucial for the challenge of detecting beyond the Standard Model physics in collider events, along with the potential for unprecedented computational efficiency in future quantum processors.
Near-term utilization of these benefits can be achieved by developing quantum-inspired algorithms for deployment in classical hardware to enable applications at the ``edge" of current scientific experiments. 
This work demonstrates the use of tensor networks for real-time anomaly detection in collider detectors.
A spaced matrix product operator (SMPO) is developed that provides sensitivity to a variety beyond the Standard Model benchmarks, and can be implemented in field programmable gate array hardware with resources and latency consistent with trigger deployment.  
The cascaded SMPO architecture is introduced as an SMPO variation that affords greater flexibility and efficiency in ways that are key to edge applications in resource-constrained environments. 
These results reveal the benefit and near-term feasibility of deploying quantum-inspired ML in high energy colliders. 

\end{abstract}

\clearpage

%%%%%%%%%%%%%%%%%%%%%%%%%%%%%%%%%%%%%%%%%%%%%%%%%%%%%%%%%%%
%%%%%%%%%%%%%%%%%%%%    Introduction    %%%%%%%%%%%%%%%%%%%
%%%%%%%%%%%%%%%%%%%%%%%%%%%%%%%%%%%%%%%%%%%%%%%%%%%%%%%%%%%

\section{Introduction}
\label{sec:intro}
%-- motivates AD 
%A crucial goal of modern particle physics since the Higgs boson discovery has been to pin down evidence of beyond the Standard Model physics. 
%The broad and unknown nature of this task has led to the rise of machine learning (ML) techniques in searches, ranging from better classification of target signal models from Standard Model background, to model-independent anomaly detection capable of achieving sensitivity to general phase space for new physics and expanding discovery potential of multipurpose search programs at the Large Hadron Collider (LHC)~\cite{BELIS2024100091}.
%Ultimately, key physics drivers such as pursuit of very rare new physics and the need for precise measurements of the Higgs and electroweak sector require ever-larger and higher energy collision datasets~\cite{p5_2023}. 

%-- LDRD intro 
Quantum computing offers new approaches to computationally intensive problems, offering significant algorithmic speedups beyond classical methods.
Quantum machine learning (ML) in particular comprises a promising class of algorithms for the exploitation of high-dimensional correlations with potentially unprecedented efficiency~\cite{Biamonte_2017, doi:10.1126/science.abn7293}, making it especially useful for the handling of large complex datasets produced by modern scientific experiments.
However, deployment on quantum hardware remains largely theoretical due to challenges such as high noise, short decoherence timescales, and instability of large-scale qubit systems~\cite{Preskill_2018, Li2025}. % have so far constrained the impact of quantum computing in real-world, high-throughput environments where it can stand to bring the most benefit. 
While awaiting the full benefits of quantum technology, hybrid classical-quantum and quantum-inspired approaches can bridge the gap between today's state-of-the-art and the quantum technologies of tomorrow through enhanced ``quantum readiness" of scientific infrastructure.
%This can be particularly useful in the scientific domain, where QML has recently enabled accelerated protein structure modeling~\cite{Zubatyuk2025} and ... 

%As a science that relies on both inherently quantum systems and produces massive amounts of complex and highly correlated data, the field of high energy physics (HEP) stands to significantly benefit from the introduction of quantum algorithms, processors, and machine learning (ML) into its experiments. 
%Define "quantum readine

One of the most promising scientific domains for the benefits of quantum computing is in high energy physics (HEP).
%While the most natural application of quantum computers is to simulate inherently quantum systems such as quantum field theories~\cite{Feynman1982}, the use of QML for efficient high-dimensional modeling can also enable new approaches for complex collider event simulation and analysis.
%Various applications of quantum and quantum-inspired ML, such as generative~\cite{PhysRevLett.126.062001, Toledo-Marin2025} and autoencoder~\cite{Duffy2025} algorithms are already being explored in HEP.
%Furthermore, the ultra-low latency and efficiency offered by QML can solve challenges associated to the unprecedented data volumes expected at future facilities such as the High-Luminosity Large Hadron Collider~\cite{Aberle:2749422} or the Future Circular Collider (FCC)~\cite{10.3389/fphy.2022.888078}, where real-time inference can manage high background rates and limited off-detector storage and processing. 
%However, quantum algorithms for real-time operation at the detector ``edge", or at the source of data, remain nascent. 
%\tcb{ I see two problems with the pargraph above, a) it doesn't connect the generative and autoencoder application to event simulation/analysis and b) something about furthermore bothers me as if this is an afterthought, i suggest following rephrasing\newline}
While the most natural application of quantum computers is to simulate inherently quantum systems such as quantum field theories~\cite{Feynman1982}, quantum-inspired algorithms can also enable new approaches for collider event simulation and analysis~\cite{PhysRevLett.126.062001, Toledo-Marin2025, Duffy2025}. 
Further, the ultra-low latency and efficiency of quantum ML can address operational challenges at future facilities such as the High-Luminosity Large Hadron Collider (HL-LHC)~\cite{Aberle:2749422} or the Future Circular Collider~\cite{10.3389/fphy.2022.888078}, where real-time inference can manage high background rates and limited off-detector storage. However, quantum algorithms for real-time operation at the detector ``edge" (or source of data) remain nascent.

%\textbf{SA: I think we are leaning maybe too hard in the QML direction, and not saying enough about Quantum Inspired ML?}
%JG: replaced QML a few places above with quantum-inspired; I don't think it's a major distinction tbh in this context 
One opportunity to explore applications of quantum-inspired ML for HEP experimental data acquisition is the study of accelerating quantum algorithms with classical edge hardware.
Traditional CPU/GPU-based simulators lack the parallelism and real-time streaming capabilities required for detector applications at the edge.
Field programmable gate arrays (FPGAs) can act as a testbed to demonstrate real-time inference with quantum-inspired ML, and as a common element of scientific data acquisition systems, could also offer opportunities for the deployment of quantum algorithms in current and near-term experiments.
Classical real-time ML is already broadly used in HEP applications~\cite{Deiana_2022}, ranging from trigger-level implementations in FPGAs~\cite{Govorkova_2022, jiang2024machinelearningevaluationglobal, Ospanov:2022fke} to front-end processing~\cite{Guglielmo_2021}. 
Recent HEP strategic planning exercises indicate the codesign of novel ML algorithms and hardware platforms as a strategic R\&D priority~\cite{apresyan2023detectorrdneedsgeneration, doe_brn}, in particular incorporating quantum algorithms and processors~\cite{gonski2026machinelearningheterogeneousedge}.  
%\tcb{ I don't have a concrete suggestion for the paragraph above but would it read better if we reorganize? there is nothing wrong but flow seems a bit off}
%-- quantum for HEP 
%In parallel to the ML revolution, the development of quantum computing technologies has advanced to an extent that quantum readiness of modern scientific infrastructure is a key consideration in the design of future facilities. 
%While the timescale for hardware realization of large-scale and stable quantum processors remains long and indefinite, inspiration from quantum algorithms and particularly quantum machine learning can be leveraged today to enable faster and more power-efficient options for a variety of current HEP challenges. 

Tensor networks (TNs) provide a natural entry point into quantum-inspired ML at the experimental edge~\cite{ORUS2014117, tns_sim, Verstraete01032008, 10.21468/SciPostPhysCore.4.1.001, Huggins_2019, tnsforml, NIPS2016_5314b967, Liu_2023}. 
First used to efficiently represent quantum many-body states obeying area-law entanglement scaling~\cite{PhysRevB.48.10345,PhysRevLett.75.3537}, tensor networks represent high-dimensional data through factorizations into interconnected low-rank tensors, efficiently capturing correlations while mitigating computational complexity. 
Crucially, inference in TN models reduces to sequences of contractions, which are bilinear operations between pairs of tensors, requiring no nonlinear activations or normalization layers. For Matrix Product States (MPS), which are 1D representations of quantum many-body states in a high-dimensional Hilbert space, and the corresponding Matrix Product Operators (MPO), the connectivity is strictly nearest-neighbor, yielding sparse contraction graphs in which each operation involves only locally adjacent tensors. 
Together, these properties, namely purely linear arithmetic and spatially local structure, make TNs particularly conducive to hardware-level parallelism and fixed-point deployment on FPGAs, where low-latency inference demands predictable data flow and minimal control overhead.
% \tcr{Tensor networks represent high-dimensional data through factorizations into interconnected low-rank tensors, efficiently capturing correlations while reducing computational complexity to sequences of optimized linear algebra operations, making them well suited for scalable and low-latency inference.
% Emphasize linearity, no activation.}

%TNs have been broadly demonstrated to connect quantum concepts to machine learning methods such as supervised~\cite{NIPS2016_5314b967} and unsupervised~\cite{Liu_2023} learning.

The complex and highly correlated nature of particle-level collider event data motivates the exploration of TNs to efficiently encode the feature space, while their low-rank factorized structure reduces both parameter count and computational cost compared to conventional ML approaches. %Moreover, the absence of nonlinear activation functions eliminates a common source of hardware complexity, and the nearest-neighbour contraction structure of 1D TNs naturally maps to pipelined or spatially parallel FPGA architectures, enabling real-time inference within the stringent resource and latency budgets imposed by collider trigger systems.
This motivation has led to TN applications in HEP including the study of lattice gauge theories ~\cite{Montangero:2021puw}, the selection of heavy flavor quarks~\cite{Felser:2020mka}, and the selection of events containing top quarks~\cite{Araz:2021zwu}. 
TNs can also be naturally leveraged for \textit{anomaly detection}~\cite{Wang:2020dvu}, defined as the recognition of outlier or out-of-class events based solely on a learned description of the background model.
This capability is especially useful in the HEP context, where the elusive nature of beyond the Standard Model (SM) physics motivates the complementarity of existing model-specific programs with broad discovery-focused searches~\cite{BELIS2024100091}.
Combining anomaly detection algorithms with FPGA deployment enables anomaly triggers for real-time event filtering, which have been recently introduced to the ATLAS~\cite{Sugizaki:2947542} and CMS~\cite{gandrakota2025realtime} experiments.
Further, use of TNs for anomaly detection in the HEP context indicates promise for these architectures to extend the new physics phase space coverage of modern collider experiments beyond classical methods~\cite{puljak2025tensornetworkanomalydetection}.

This work describes a proof of concept for a class of TNs called Spaced Matrix Product Operators (SMPOs) to be implemented in FPGAs for real-time anomaly detection at collider experiments.
% \tcr{SMPOs are specialized forms of MPOs that can act on MPOs JG: this reads a little weird to me; maybe "specialized forms of MPOs that can also act on MPOs"?}, meant to represent states in the 1D Hilbert space. 
SMPOs are specialized forms of MPOs that, like MPOs, can act on MPSs, but differ from standard MPOs in that they reduce dimensionality of the input MPS through ``spacing" of outputs.
Further, SMPOs can be trained in an unsupervised way, making them applicable for anomaly detection applications~\cite{Wang:2020dvu}.
%\tcr{ SA: Sorry, just reading the sentence above, but it seems misplaced in the current place. We go straight from defining SMPO to saying that SMPOs scale linearly with size (not sure what that means) and hence are useful for AD?}
%A challenge with SMPO deployment on FPGAs is the tight constraint on on-device resources, which limits the size of tensor calculations that can be performed. 
Bringing SMPOs to real-time applications requires dedicated work to simulate, optimize, and implement TNs in FPGA platforms; initial efforts in HEP have focused on the task of heavy flavor jet tagging~\cite{borella2024ultralowlatencyquantuminspiredmachine, coppi2026tensornetworkmodelslowlatency}. 
To further enable FPGA deployment of TN technology, we introduce the \textbf{cascaded SMPO (CSMPO)}, a
SMPO architecture refactoring that, in a resource-constrained environment, provides similar learning capacity and performance for a fraction of the computational power needed for inference. 
Both traditional SMPO and CSMPO implementations presented here can meet the performance and resource/latency requirements expected at future collider trigger systems, opening the door to the advent of quantum-inspired ML in HEP data pipelines. 
%Ours is the first AD algorithm demonstrated to be feasible for FPGA implementation, opening the door to a new era of model-independent real-time data acquisition at high data rate experiments. 

%%%%%%%%%%%%%%%%%%%%%%%%%%%%%%%%%%%%%%%%%%%%%%%%%%%%%%%%%%%
%%%%%%%%%%%%%%%%%%%%    Methods    %%%%%%%%%%%%%%%%%
%%%%%%%%%%%%%%%%%%%%%%%%%%%%%%%%%%%%%%%%%%%%%%%%%%%%%%%%%%%

\section{Methodology}
\label{sec:methods}

%--------------------------------------------------------
\subsection{Input Modeling}

\subsubsection{Samples}

The TN anomaly detection models are developed using simulated events of proton-proton collisions at the Large Hadron Collider~\cite{Govorkova_2022, govorkova2021lhcphysicsdatasetunsupervised}. 
Events are pre-filtered to require the presence of at least one energetic electron or muon. 
This dataset includes five simulated processes: a background consisting of multijet events arising from quantum chromodynamics (QCD), and four beyond the Standard Model signals: 
\begin{enumerate} 
\item A neutral scalar boson $A$ decaying via two $Z$ bosons to a four-lepton final state, $A\to4\ell$;
\item A leptoquark (LQ) decaying to a $b$ quark and $\tau$ lepton, $LQ\to b\tau$;
\item A charged scalar boson $h^\pm$ decaying to a $\tau$ lepton and neutrino $\nu$, $h^\pm\to\tau\nu$;
\item A neutral scalar boson $h^0$ decaying to two $\tau$ leptons, $h^0\to\tau\tau$.
\end{enumerate}
The anomaly detection capability of the method can be determined by the ability of the model to distinguish all four signals from the background, despite their varying characteristics.

Events are modeled by 57 variables from 19 particles: the three-vectors (transverse momentum $p_\mathrm{T}$, pseudorapidity $\eta$, and azimuthal angle $\phi$) of the leading ten jets ($j$), four electrons ($e$), four muons ($\mu$), and the missing energy ($E_\mathrm{T}^\mathrm{miss}$). Since the missing energy is only measured in the transverse plane, its pseudorapidity is set to 0 for all events. If any of these particles aren't reconstructed for an event, their three-vectors are zero-padded.

%\tcb{ a naive question, is there a motivation to only use these particles? a reader might think why photon is not considered? }

\subsubsection{Input Embedding}

Every event is embedded into a Matrix Product State (MPS), a one-dimensional linear chain of tensors. 
The embedding is chosen such that one tensor site represents the kinematics of one reconstructed particle, making a total of 19 tensor sites. The four-vector kinematics of the particles map to the tensor sites through a simple pre-processing to bring the tensor element values to similar ranges ($O(0,1)$), and also to prevent a tensor norm collapse from the zero-padded particles. %Equation~\ref{eq:preprocessing} shows 
The values of the scaling used for each variable for a particle at site $i$ are: 

\begin{equation}
   \begin{aligned}
      x_{i,1} &= \frac{p_{\mathrm{T},i}}{p_\mathrm{T,ref}}\\
      x_{i,2} &= \frac{\eta_i+5}{10} \\
      x_{i,3} &= \frac{\phi_i+\pi}{2\pi}\\
   \end{aligned}
   \label{eq:preprocessing}
\end{equation}
Here $p_\mathrm{T,ref}$ is set to 2500 GeV for jets, 800 GeV for muons, and 1200 GeV for electrons and $E_\mathrm{T}^\mathrm{miss}$, chosen approximately based on the $p_\mathrm{T}$ spread of the different particle classes. 

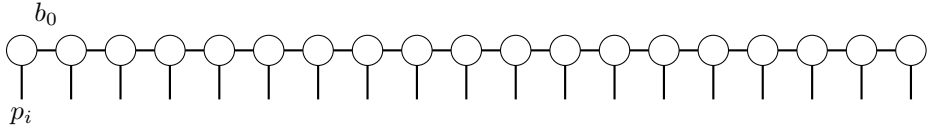
\begin{figure}[!ht]
   \centering
   \begin{tikzpicture}[scale=.65]
      %% Create a matrix site
      \tikzset
      {
      site/.style={circle, inner sep=0pt, outer sep=0pt, minimum width=.4cm},
      selected site/.style = {site, fill=red!30},
      % nedge/.style = {-}
      }
      %% Sites
      \node[draw, site] (s0) at (0,0) {};
      \node[draw, site] (s1) at (1,0) {};
      \node[draw, site] (s2) at (2,0) {};
      \node[draw, site] (s3) at (3,0) {};
      \node[draw, site] (s4) at (4,0) {};
      \node[draw, site] (s5) at (5,0) {};
      \node[draw, site] (s6) at (6,0) {};
      \node[draw, site] (s7) at (7,0) {};
      \node[draw, site] (s8) at (8,0) {};
      \node[draw, site] (s9) at (9,0) {};
      \node[draw, site] (s10) at (10,0) {};
      \node[draw, site] (s11) at (11,0) {};
      \node[draw, site] (s12) at (12,0) {};
      \node[draw, site] (s13) at (13,0) {};
      \node[draw, site] (s14) at (14,0) {};
      \node[draw, site] (s15) at (15,0) {};
      \node[draw, site] (s16) at (16,0) {};
      \node[draw, site] (s17) at (17,0) {};
      \node[draw, site] (s18) at (18,0) {};
      %% Bonds
      \draw [thick] (s0) -- (s1) node[midway, above=0.2cm] {$b_0$} -- (s2) -- (s3) -- (s4) -- (s5) -- (s6) -- (s7) -- (s8) -- (s9) -- (s10) -- (s11) -- (s12) -- (s13) -- (s14) -- (s15) -- (s16) -- (s17) -- (s18)
      %% Outgoing physical indices
      (s0) -- ++(0,-1) node[midway, below=0.2cm] {$p_i$}
      (s1) -- ++(0,-1)
      (s2) -- ++(0,-1)
      (s3) -- ++(0,-1)
      (s4) -- ++(0,-1)
      (s5) -- ++(0,-1)
      (s6) -- ++(0,-1)
      (s7) -- ++(0,-1)
      (s8) -- ++(0,-1)
      (s9) -- ++(0,-1)
      (s10) -- ++(0,-1)
      (s11) -- ++(0,-1)
      (s12) -- ++(0,-1)
      (s13) -- ++(0,-1)
      (s14) -- ++(0,-1)
      (s15) -- ++(0,-1)
      (s16) -- ++(0,-1)
      (s17) -- ++(0,-1)
      (s18) -- ++(0,-1);
   \end{tikzpicture}
   \caption{Embedded particle MPS that represents the input to the SMPO and CSMPO models. $p_i$ refers to the physical dimension, here with a value of 3, and $b_0$ refers to the trivial bond dimension with value 1.}
   \label{fig:mps}
\end{figure}

The event MPS $\mathbf{X}$ is hence constructed as a tensor product of all the individual sites $\mathbf{x}_i$ as:
\begin{equation}
   \mathbf{X} = \frac{1}{\Gamma}\bigotimes_{i=1}^{19} \mathbf{x}_i,  \ \ \ \ \ \ \ \ 
   \Gamma=\left({\displaystyle \prod_{i=1}^{19}} \|\mathbf{x_i}\|\right)^{1/19}
   \label{eq:input_MPS}
\end{equation}
where $\Gamma$ is an MPS normalization factor.
In tensor diagram notation, the embedded event is represented as in Figure~\ref{fig:mps}. 

MPSs are described by two key hyperparameters: the physical dimension, i.e. the length of each site, and the bond dimension, i.e. the size of interaction between neighboring sites. 
Tensor networks need higher bond dimensions to capture long range correlations in learned systems~\cite{RevModPhys.82.277}. 
However, the cost of tensor mathematical operations scale with increasing bond dimensions (see discussion in Appendix~\ref{app:mac}). 
Hence, a trade-off in bond dimension size needs to be chosen to balance performance with resource cost and must be optimized throughout the architecture.
For the embedding MPS, the physical dimension ($p_i$) of the MPS is 3, dictated by the three-vector representing each particle, with a trivial bond dimension ($b_0$) of 1 (as each site is a tensor of rank 1).

\subsubsection{Input Ordering}

The features that make up the input MPS are ordered based on a statistical equivalent of Quantum Mutual Information (QMI)~\cite{PhysRevLett.79.5194,Nielsen_Chuang_2010}, which measures the degree of correlations between pairwise sites. The QMI between two sites $i$ and $j$ is defined as:
\begin{equation}
   \mathrm{QMI}_{ij} = S(\rho_i)+S(\rho_j)-S(\rho_{ij}),
\end{equation}
where $S(\rho)=-\mathrm{Tr}(\rho\log\rho)$ is the von Neumann entropy and $\rho_i$ is the $3\times3$ density matrix for site $i$ and $\rho_{ij}$ is the $9\times9$ two-site density matrix for sites $i$ and $j$, defined as follows where $n$ runs over the training events in the input dataset.

\begin{equation}
   \begin{aligned}
      \rho_i &= \frac{1}{N}\sum_{n=1}^{N} \mathbf{x}_i^n \mathbf{x}_i^{n\mathrm{T}} \\
      \rho_{ij} &= \frac{1}{N}\sum_{n=1}^{N} (\mathbf{x}_i^n \otimes \mathbf{x}_j^n)(\mathbf{x}_i^n \otimes \mathbf{x}_j^n)^\mathrm{T}
   \end{aligned}
\end{equation}

\begin{figure}[!htbp]
\centering
    \subfloat[Normal Ordering\label{fig:qmi_normal}]{\includegraphics[width=0.45\linewidth]{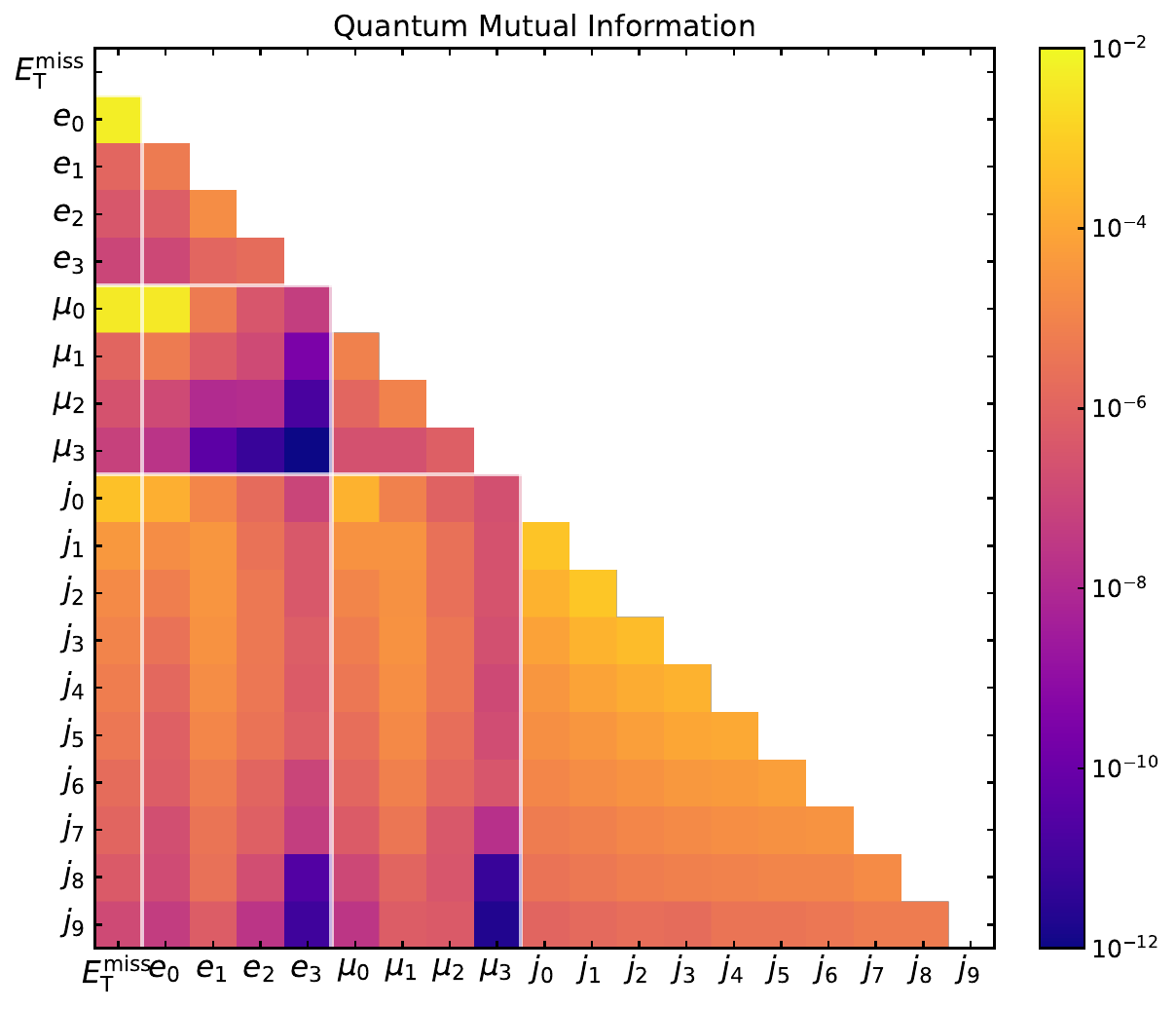}}
    \subfloat[Spectral Ordering\label{fig:qmi_spectral}]{\includegraphics[width=0.45\linewidth]{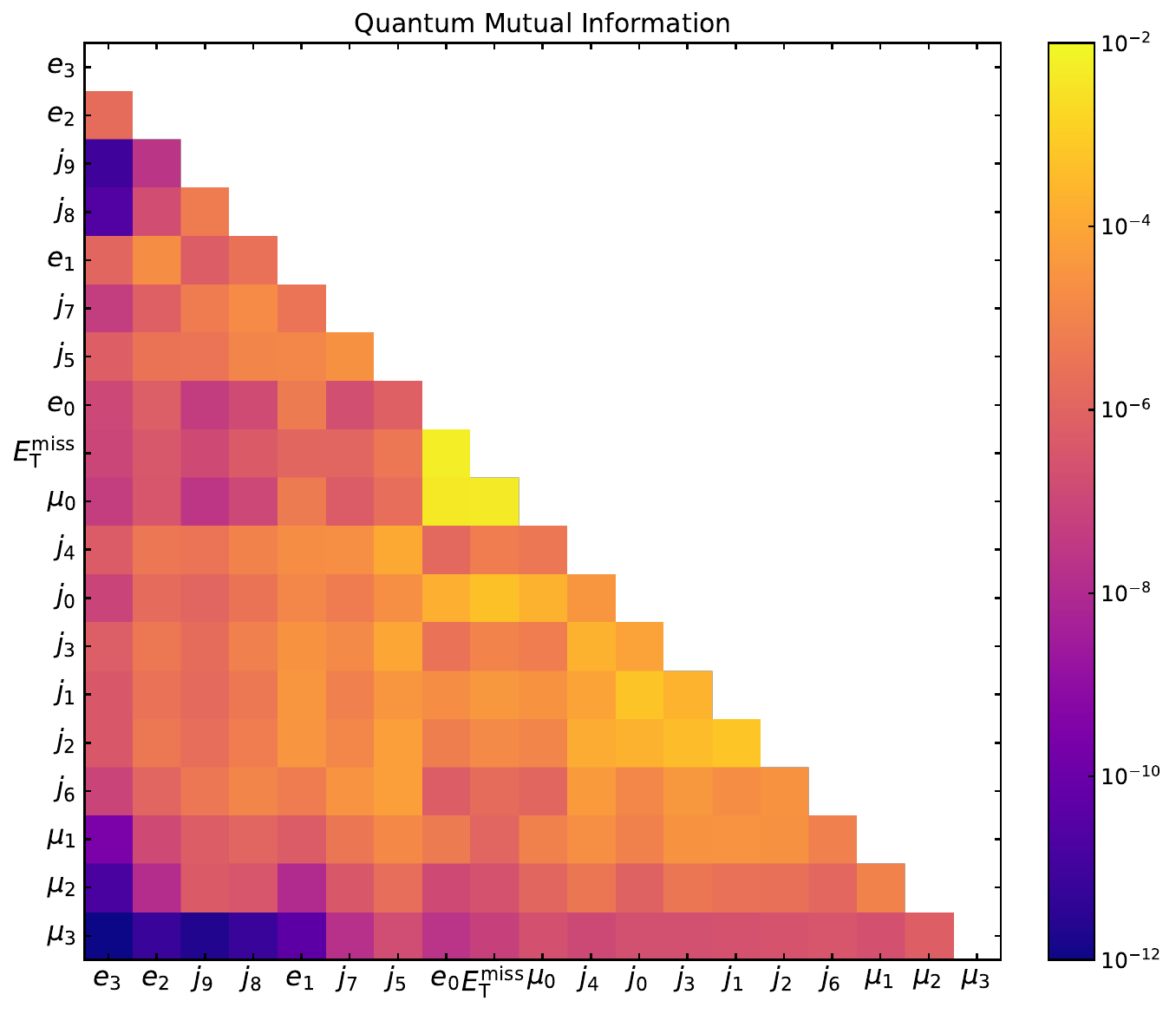}}
    \caption{Quantum Mutual Information between sites of the input MPS in the SMPO architecture, comparing ordering based on particle type and $p_\mathrm{T}$ (left) and spectral ordering (right). The use of spectral ordering ensures particles with high QMI are near each other in the SMPO structure, allowing for reduced bond dimensions to capture correlations between inputs.}
    \label{fig:qmi}
\end{figure}

A higher QMI points to a higher degree of mutual information between two particles which is more effectively learned by a tensor network model if kept next to each other.
A spectral ordering~\cite{Acharya2022QubitSeriation} algorithm is used to order the sites in the MPS, ensuring that the sites with highest QMI are clustered together in the middle of the MPS. 
By ensuring that particles with the highest QMI are kept together, the tensor network model applied on the input dataset can be trained with smaller bond dimensions than with a particle-based ordering, capturing relevant correlations while reducing the computational overhead. 
Figure~\ref{fig:qmi} shows the QMI of the sites comparing two choices of ordering: one grouped by particle type and in each group ordered by descending $p_\mathrm{T}$, and one using a spectral ordering. Since the dataset requires at least one energetic $e$ or $\mu$ in each event, the $E_\mathrm{T}^\mathrm{miss}$ measurement is expected to be correlated to the leading $e/\mu$ energy scale, hence the high QMI between them. Furthermore, since the underlying physics of the training dataset is multijets from QCD, the leading jets are expected to have higher QMI with the other dominant features in this ordering.

%\tcb{For figure 2 we either have to explain what normal ordering means or even name as random ordering or not show normal ordering. Since there is no discussion of normal ordering it is confusing. \newline We might also have to add some description to guide the user on how to read the spectral ordered plot.\newline Sagar do you have intuition on why some objects have higher QMI like mu0/EtMiss, we might get asked during review}

\subsection{Architectures}

\subsubsection{SMPO}

SMPOs are well-suited to anomaly detection applications as they provide a dimensionality reduction when acting on an $N-$site MPS and returning an $M-$site MPS, where $M<N$. 
This compression enables the model to learn salient features of the training dataset without class labels. 
Typically, a constant spacing is used between sites with physical output legs. 
In this study, a most aggressive form of an SMPO is used for reducing the dimensionality of a 19-site MPS to a trivial MPS of one-site, i.e. a vector. 

\begin{figure}[!ht]
   \centering
   \begin{tikzpicture}[scale=.65]
      %% Create a matrix site
      \tikzset
      {
      site/.style={circle, inner sep=0pt, outer sep=0pt, minimum width=.4cm},
      selected site/.style = {site, fill=red!30},
      % nedge/.style = {-}
      }
      %% Sites
      \node[draw, site] (s0) at (0,0) {};
      \node[draw, site] (s1) at (1,0) {};
      \node[draw, site] (s2) at (2,0) {};
      \node[draw, site] (s3) at (3,0) {};
      \node[draw, site] (s4) at (4,0) {};
      \node[draw, site] (s5) at (5,0) {};
      \node[draw, site] (s6) at (6,0) {};
      \node[draw, site] (s7) at (7,0) {};
      \node[draw, site] (s8) at (8,0) {};
      \node[draw, selected site] (s9) at (9,0) {};
      \node[draw, site] (s10) at (10,0) {};
      \node[draw, site] (s11) at (11,0) {};
      \node[draw, site] (s12) at (12,0) {};
      \node[draw, site] (s13) at (13,0) {};
      \node[draw, site] (s14) at (14,0) {};
      \node[draw, site] (s15) at (15,0) {};
      \node[draw, site] (s16) at (16,0) {};
      \node[draw, site] (s17) at (17,0) {};
      \node[draw, site] (s18) at (18,0) {};
      %% Bonds
      \draw [thick] (s0) -- (s1) node[midway, above] {$b$} -- (s2) -- (s3) -- (s4) -- (s5) -- (s6) -- (s7) -- (s8) -- (s9) -- (s10) -- (s11) -- (s12) -- (s13) -- (s14) -- (s15) -- (s16) -- (s17) -- (s18)
      %% Incoming physical indices
      (s0) -- ++(0,1) node[midway, above=0.2cm] {$p_i$}
      (s1) -- ++(0,1)
      (s2) -- ++(0,1)
      (s3) -- ++(0,1)
      (s4) -- ++(0,1)
      (s5) -- ++(0,1)
      (s6) -- ++(0,1)
      (s7) -- ++(0,1)
      (s8) -- ++(0,1)
      (s9) -- ++(0,1)
      (s10) -- ++(0,1)
      (s11) -- ++(0,1)
      (s12) -- ++(0,1)
      (s13) -- ++(0,1)
      (s14) -- ++(0,1)
      (s15) -- ++(0,1)
      (s16) -- ++(0,1)
      (s17) -- ++(0,1)
      (s18) -- ++(0,1)
      %% Middle outline
      (s9) -- ++(0,-1) node[midway, below=0.2cm] {$p_o$};
   \end{tikzpicture}
   \caption{Diagram of the SMPO model used in this work, where $b$ refers to the bond dimension and has a value of 4, $p_i$ refers to input physical dimension with a value of 3, and $p_o$ refers to the physical output dimension with a value of 3. The shaded node refers to the site in the SMPO with a physical output leg.}
   \label{fig:smpo}
\end{figure}
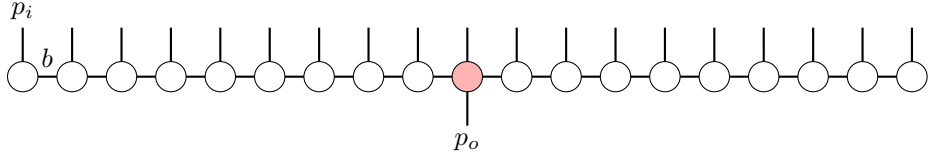

Figure~\ref{fig:smpo} shows the architecture of the SMPO developed for the anomaly triggering task. 
The input physical dimension $p_i$ is the same as the physical dimension of the input MPS, i.e. 3. 
The output physical dimension $p_o$ and the bond dimension $b$ are optimized for task performance. An anomaly detection task requires a model to learn optimally about low and high-level correlations in the underlying dataset without learning about extrapolations to out-of-domain anomalous events. Hence, the optimization of these two model parameters are done to only reduce model size while retaining learning capacity over the training dataset (described further in Section~\ref{sec:result}). 
%which is learned from the TPR metric across the four benchmark signals, as defined in . 
The optimal bond dimension $b$ is found to be 4, and $p_o$ to be 3. 
This model architecture is described as 19$\rightarrow$1, referring to the input and output sizes. 
%The bond dimension of the SMPO is optimized for performance on anomalous signals. While a large bond dimension can capture long-range correlations in the underlying dataset, it can also relax the strong bottleneck required for anomaly detection capabilities. On the other hand, a small bond dimension can fail to fully capture the correlations across sites, and hence not providing enough learning power on both the trained background and the anomalous signals. The optimized bond dimension $b$ that maximises anomaly detection performance on benchmark signals is found to be 4.

% Within TNs, there are two classes of architectures \tcr{commonly used? Citations here? Why MPO, SMPO?}: Matrix Product Operator (MPO), where each tensor has two uncontracted indices and the neighbors have contracted indices; and Spaced MPOs (SMPOs), where some outlines are removed (typically periodic, giving an integer input-output ratio).
% Corner vs. center out-line. 

\subsubsection{Cascaded SMPO}

The CSMPO is a consecutive application of multiple SMPOs for serialized dimensionality reduction. 
Since SMPOs are linear mathematical operators, a consecutive application of two such operators is itself an SMPO, and the effective bond dimension of the composite is the product of the individual bond dimensions:
\begin{equation}
   \mathrm{SMPO}_{19\to M} (b_1) \times \mathrm{SMPO}_{M\to 1} (b_2) \subset \mathrm{SMPO}_{19\to 1} (b1 \times b2)
\end{equation}
%\tcg{\sout{Despite the high level mathematical equivalence, the structure of the CSMPO leads to fundamental differences in learning capacity, namely more flexible hyperparameter space when trained for the same anomaly detection task.}
The inclusion is strict: the cascade structure imposes a factorization constraint on the effective weight tensors, so not every single-layer SMPO admits a cascade decomposition at reduced bond dimension. 
However, this restricted parameterization ultimately allows for a more flexible hyperparameter space for optimization.
The spacing of the first SMPO layer and the output physical dimension of the intermediate MPS state add degrees of freedom while reducing the number trainable parameters as compared to the equivalent SMPO, and hence are hypothesized to offer similar expressivity from a smaller model. 
In addition, the CSMPO model poses an opportunity for reduced resource use coming from the bond dimension being split across two model layers, instead of a larger bond dimension used in a single layer, since the cost of mathematical operations scale quadratically or cubically with the bond dimension (see discussion in Appendix~\ref{app:mac}).
% \tcr{JG: Can you add one more sentence here roughly explaining why the single layer high bond dimension calculations are so costly? could forward ref to the contractions maybe}

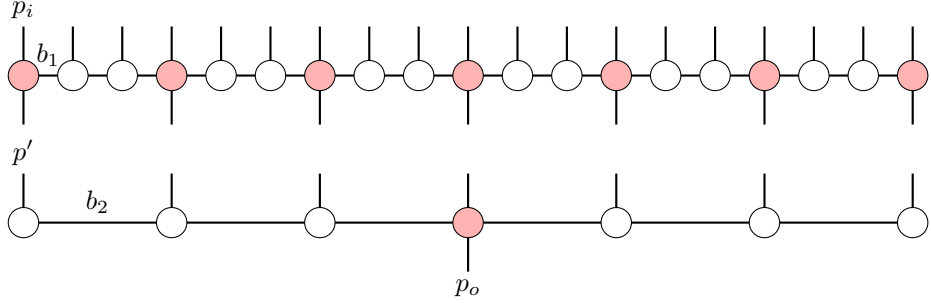
\begin{figure}[!ht]
   \centering
   %\subfloat[%CSMPO$_{19\to7\to1}$
   %\label{fig:csmpo_19_7_1}]
   {
      \begin{tikzpicture}[scale=.65]
         \tikzset
         {
         site/.style={circle, inner sep=0pt, outer sep=0pt, minimum width=.4cm},
         selected site/.style = {site, fill=red!30},
         % nedge/.style = {-}
         }
         %% First layer
         \node[draw, selected site] (f0) at (0,0) {};
         \node[draw, site] (f1) at (1,0) {};
         \node[draw, site] (f2) at (2,0) {};
         \node[draw, selected site] (f3) at (3,0) {};
         \node[draw, site] (f4) at (4,0) {};
         \node[draw, site] (f5) at (5,0) {};
         \node[draw, selected site] (f6) at (6,0) {};
         \node[draw, site] (f7) at (7,0) {};
         \node[draw, site] (f8) at (8,0) {};
         \node[draw, selected site] (f9) at (9,0) {};
         \node[draw, site] (f10) at (10,0) {};
         \node[draw, site] (f11) at (11,0) {};
         \node[draw, selected site] (f12) at (12,0) {};
         \node[draw, site] (f13) at (13,0) {};
         \node[draw, site] (f14) at (14,0) {};
         \node[draw, selected site] (f15) at (15,0) {};
         \node[draw, site] (f16) at (16,0) {};
         \node[draw, site] (f17) at (17,0) {};
         \node[draw, selected site] (f18) at (18,0) {};
         %% Second layer
         \node[draw, site] (s0) at (0,-3) {};
         \node[draw, site] (s1) at (3,-3) {};
         \node[draw, site] (s2) at (6,-3) {};
         \node[draw, selected site] (s3) at (9,-3) {};
         \node[draw, site] (s4) at (12,-3) {};
         \node[draw, site] (s5) at (15,-3) {};
         \node[draw, site] (s6) at (18,-3) {};
         %% Layer 1 bonds
         \draw [thick] (f0) -- (f1) node[midway, above] {$b_1$} -- (f2) -- (f3) -- (f4) -- (f5) -- (f6) -- (f7) -- (f8) -- (f9) -- (f10) -- (f11) -- (f12) -- (f13) -- (f14) -- (f15) -- (f16) -- (f17) -- (f18)
         %% Layer 1 incoming physical indices
         (f0) -- ++(0,1) node[midway, above=0.2cm] {$p_i$}
         (f1) -- ++(0,1)
         (f2) -- ++(0,1)
         (f3) -- ++(0,1)
         (f4) -- ++(0,1)
         (f5) -- ++(0,1)
         (f6) -- ++(0,1)
         (f7) -- ++(0,1)
         (f8) -- ++(0,1)
         (f9) -- ++(0,1)
         (f10) -- ++(0,1)
         (f11) -- ++(0,1)
         (f12) -- ++(0,1)
         (f13) -- ++(0,1)
         (f14) -- ++(0,1)
         (f15) -- ++(0,1)
         (f16) -- ++(0,1)
         (f17) -- ++(0,1)
         (f18) -- ++(0,1)
         %% Layer 1 outgoing physical indices
         (f0) -- ++(0,-1) node[midway, below=0.3cm] {$p'$}
         (f3) -- ++(0,-1)
         (f6) -- ++(0,-1)
         (f9) -- ++(0,-1)
         (f12) -- ++(0,-1)
         (f15) -- ++(0,-1)
         (f18) -- ++(0,-1);
         %% Layer 2 bonds
         \draw [thick] (s0) -- (s1) node[midway, above] {$b_2$} -- (s2) -- (s3) -- (s4) -- (s5) -- (s6)
         (s0) -- ++(0,1)
         (s1) -- ++(0,1)
         (s2) -- ++(0,1)
         (s3) -- ++(0,1)
         (s4) -- ++(0,1)
         (s5) -- ++(0,1)
         (s6) -- ++(0,1)
         %% Middle outline
         (s3) -- ++(0,-1) node[midway, below=0.2cm] {$p_o$};
      \end{tikzpicture}
   } \\
   \caption{Diagram of the CSMPO model used in this work, where $b_1$ and $b_2$ refer to the bond dimensions of the first and second layers respectively with values $b_1$ = $b_2$ = 2, $p_i$ refers to input physical dimension with a value of 3, $p'$ refers to physical dimension of the intermediate MPS created between the two layers with a value of 3, and $p_o$ refers to the output physical dimension with a value of 3. The shaded nodes refer to the sites in the CSMPO layers with physical output legs.}
   \label{fig:csmpo}
\end{figure}

Figure~\ref{fig:csmpo} provides a diagram of the CSMPO architecture optimized for the anomaly detection task.
The first layer is an SMPO reducing the MPS length from 19 to 7 sites through a spacing of 3, and the second layer reduces it further from 7 to 1. 
The input physical dimension $p_i$, output physical dimension $p_o$, and the output physical dimension of the MPS created between the two layers $p'$ are all 3. 
Compared to the SMPO, $p'$ is an additional optimizable parameter in the CSMPO. %, and the final value was chosen based on anomaly detection performance.
The bond dimensions of the two layers $b_1$ and $b_2$ are set to $b_1=b_2=2$, such that the tensor architecture is mathematically equivalent to the SMPO defined above, which has a bond dimension of $b=4=b_1 \times b_2$.
Similar to the SMPO, this model structure is referred to by its dimensionality as CSMPO$_{19\to7\to1}$.
The intermediate layer provides a reduction in the number of trainable parameters due to a high sparsity in the first layer coming from the spacing of 3, where the extent of the model compression arises from the amount of spacing at each layer.
%~\tcr{SA: This is a nuanced point since if we had two layers of bond-dimension of two each, the numbers of parameters would have been the same as the SMPO if the first layer has spacing of 1 (i.e. every site has an outline). This is also why $19\to2\to1$ has even fewer training parameters, since it has a spacing of 18.}
%\tcr{due to some bond dimension thing? Sagar please fill in}, 
This leads to a total trainable parameter count of 456 for the CSMPO, an approximate 50\% reduction compared to 936 for the SMPO. 
   %\item CSMPO$_{19\to2\to1}$ -~\ref{fig:csmpo_19_2_1} shows the model architecture. The first layer is an SMPO reducing the MPS length from 19 to 2 sites through a spacing of 18, and the second layer reduces it further from 2 to 1. The physical dimensions of this model are the same as that of CSMPO$_{19\to7\to1}$.

%\tcr{SA: Should we say anything about number of trainable parameters in the models?}
%\tcb{this all looks great, i think the only thing missing is to say that CSMPO physical dimensions are also all optimized to be 3 instead of stating they are 3}

%--------------------------------------------------------
\subsection{Training Details}

%The bond dimension of the SMPO model is determined via a singular value decomposition (SVD) analysis of the intermediate MPS representations.
%The singular value spectrum is computed across all sites, and the minimal bond dimension required to retain 90\% of the cumulative singular value weight is selected, yielding a bond dimension of 4.
%Sagar: The above is not true. We can't do SVD on an MPS with just 1 site. We optimize the bond-dim for performance.

TN models are implemented using \texttt{tn4ml}~\cite{puljak2025tn4mltensornetworktraining}, a software library for building ML pipelines with TNs.
Both SMPO and CSMPO models are trained on background events using a Pseudo-Huber loss function~\cite{huber1964robust} applied to $\|\mathbf{MPS}(x)\|^2$, the squared norm of the final MPS output for a given input event $\mathbf{x}$.
The loss $\mathcal{L}$ is defined as
\begin{equation}
   \begin{aligned}
      \mathcal{L}(\mathbf{x}|\mu,\delta) ={} & \delta^2\left(\sqrt{1 + \left(\frac{\|\mathbf{MPS}(\mathbf{x})\|^2 - \mu}{\delta}\right)^2} - 1\right) \\
      & + \begin{cases}
         \log^2\!\left(\frac{\|\mathbf{MPS}(\mathbf{x})\|^2}{\mu}\right) & \text{if }\|\mathbf{MPS}(\mathbf{x})\|^2 < 1 \\
         0 & \text{otherwise}
      \end{cases}
   \end{aligned}
\end{equation}
where $\mu$ is the target squared norm and $\delta$ is the Huber smoothing parameter. The second term in the loss is a weak collapse-prevention penalty, active only when the norm approaches zero. 
The value of the target squared norm $\mu$ is approximately chosen to capture the full spread of the $\|\mathbf{MPS}\|^2$ distribution to ensure the model can retain the relevant information for making an anomaly decision (see Figure~\ref{fig:smpo_mps}).

Each model is independently optimized over the hyperparameters $\mu$ and $\delta$.
The Huber loss behaves quadratically for $|\|\mathbf{MPS}\|^2-\mu|<\delta$ and only linearly for $|\|\mathbf{MPS}\|^2-\mu|>\delta$. 
This reduction in outlier pull allows for defining an anomaly threshold consistent with the needs of the application. %, such as the anomaly triggering task in this work. 
%A larger $\delta$ value captures more outliers increasing the AUC, whereas a smaller $\delta$ trades off some AUC for increased TPR, %where AUC and TPR are defined in Section~\ref{sec:result}.
%JG: I think this AUC vs. TPR point is kind of vague anyway and really messed up the flow without defining these quantities first, so lets just get rid of it 
Optimizing for the triggering application, where performance is assessed in the very low background acceptance range, leads to values of  $(\mu, \delta) = (50, 25)$ and $(50, 15)$ for the SMPO and CSMPO, respectively.

As the models are trained unsupervised, the training dataset comprises only QCD background events. 
The full QCD sample is split into training (70\%), validation (5\%), and test (25\%) subsets, corresponding to 2.8 million training events, 200,000 validation events, and 1 million test events.
Both models are optimized using Adam~\cite{kingma2017adammethodstochasticoptimization} with a batch size of 2048, and learning rates of $4\times10^{-3}$ and $10^{-2}$ for SMPO and CSMPO respectively.
Training proceeds for up to 200 epochs with early stopping based on validation loss, with a patience of 50 epochs and a minimum improvement threshold of $10^{-4}$, restoring the best-performing checkpoint upon convergence.

On evaluation of new events, the anomaly score is defined as the absolute deviation of the $\|\mathbf{MPS}\|^2$ value from the median of the background test sample:
\begin{equation}
    \mathcal{S}(\mathbf{x}) = \left|\,\|\mathbf{MPS}(\mathbf{x})\|^2 - \text{median}_{\text{bkg}}\!\left(\|\mathbf{MPS}\|^2\right)\right|.
\end{equation}

%%%%%%%%%%%%%%%%%%%%%%%%%%%%%%%%%%%%%%%%%%%%%%%%%%%%%%%%%%%
%%%%%%%%%%%%%%%%%%%%    Results    %%%%%%%%%%%%%%%%%
%%%%%%%%%%%%%%%%%%%%%%%%%%%%%%%%%%%%%%%%%%%%%%%%%%%%%%%%%%%

\section{Results}
\label{sec:result}
Results are provided for the SMPO and CSMPO models in terms of two key features: performance at the anomaly detection task, and computational burden of model inference to demonstrate the viability of real-time deployment. 

Anomaly detection performance is evaluated based on the model receiver operating characteristic  (ROC) curves, which show signal efficiency (true positive rate, TPR) as a function of the background efficiency (false positive rate, FPR). 
The ROC area-under-curve (AUC) is used as a metric for the ability of the SMPO anomaly score to discriminate the four test signals from background. 
Additionally, signal acceptance at very low background efficiencies of $10^{-5}$ FPR is evaluated for each signal process (referred to simply as ``TPR" in the following discussion). 
As trigger application requires to be able to efficiently select signal processes at very high background rates, the performance of the model in the extreme FPR range is critical for model design and optimization. 

The computational overhead for each model is assessed by the number of model parameters and the number of multiply-and-accumulate (MAC) operations needed for inference.
A complete explanation of the MAC calculation for each model can be found in Appendix~\ref{app:mac}.
An additional means of assessing model complexity is achieved through the synthesis of the algorithm to FPGA block resources, namely look-up table (LUTs), digital signal processors (DSPs), and flip-flops (FFs), as well as inference latency in an FPGA implementation.

%------------------------------------------------------
\subsection{Model Performance}

\begin{figure}[tbh]
\centering
\includegraphics[width=0.85\textwidth]{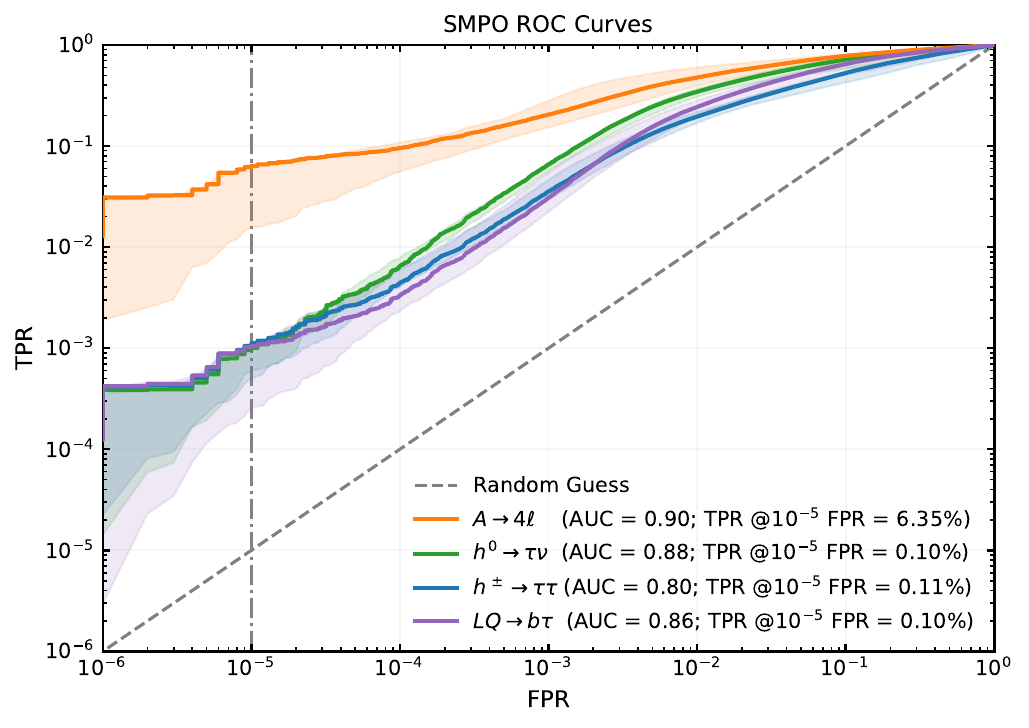}
\caption{\label{fig:smpo_perf} ROC curve for the SMPO model, indicating AUC and TPR for all four test signals. The error band reflects the ensembled performance of ten identical model trainings with different random seeds. The solid line indicates the best performing model in the ensemble.}
\end{figure}

Figure~\ref{fig:smpo_perf} shows the ROC curve of the SMPO anomaly score for all four test signals. 
To mitigate the influence of stochasticity, an error band is shown that reflects the performance of an ensemble of ten models, each trained under identical circumstances except with different random seeds. The model with the highest TPR is chosen as the standard, and numbers are reported for this model.

The AUC spans 0.80-0.90 across the four signals, indicating good overall discrimination from an unsupervised training with no explicit use of signal model features.
While TPR for the other signals is $\sim0.10\%$, the model achieves a TPR of 6.35\% for $A\to4\ell$, a very large acceptance even at extremely high background rejection. A four-lepton event is kinematically very distinct from QCD multijet background. The high discrimination power displayed by the SMPO model attests to its topological anomaly detection capability, especially for trigger applications which require extremely small FPRs. These results are generally consistent with, and for some signals outperform, state-of-the-art methods that leverage traditional ML~\cite{Govorkova_2022}.

\begin{figure}[tbh]
\centering
\includegraphics[width=0.85\textwidth]{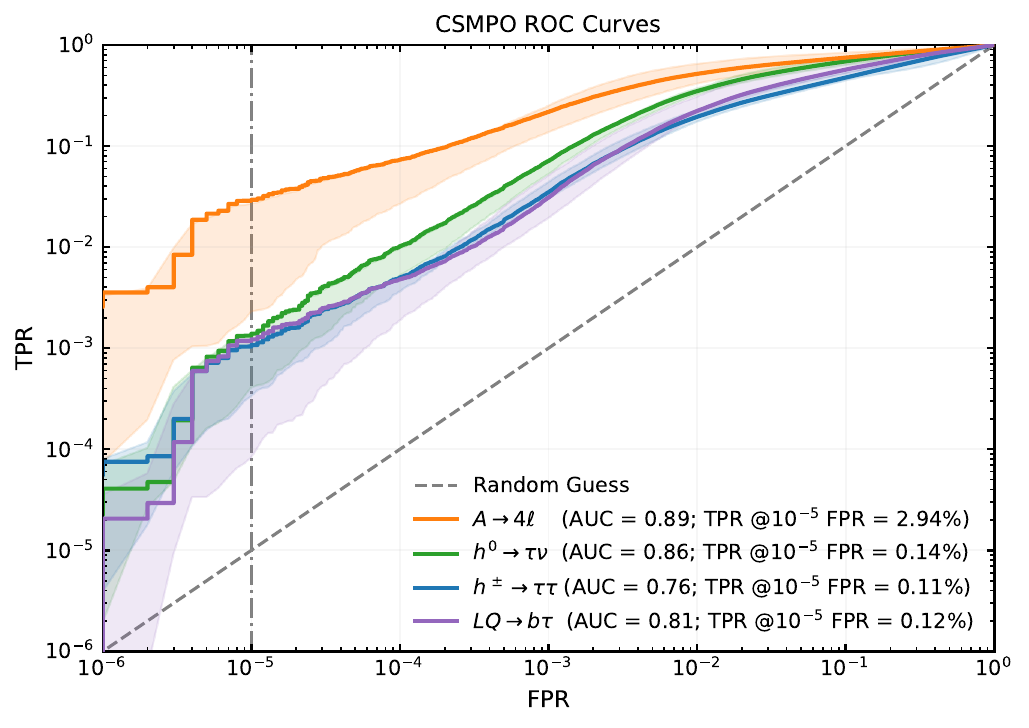}
\caption{\label{fig:csmpo_perf} ROC curve for the CSMPO model, indicating AUC and TPR for all four test signals. The error band reflects the ensembled performance of ten identical model trainings with different random seeds. The solid line indicates the best performing model in the ensemble.}
\end{figure}

Figure~\ref{fig:csmpo_perf} compares the four signal ROC curves for the CSMPO model. 
%\tcg{\sout{For each signal process, the CSMPO model exhibits improved performance relative to the SMPO in terms of AUC.
%This improvement is observed consistently across all four signal processes. 
%However, its performance in TPR is reduced considerably, most noticeably for $A\to4\ell$, going from 6.35\% to 2.94\%.}
Comparisons to the SMPO model indicate roughly similar performance overall, with some improvements and degradations across the signals and performance metrics. 
For each signal process, the CSMPO model exhibits a marginal degradation in performance relative to the SMPO in terms of AUC. 
Its performance in TPR varies across signals, improving marginally over the SMPO for the $h^0\to\tau\nu$ and $LQ\to b\tau$ signals, while reducing considerably for $A\to4\ell$. 
%going from 6.35\% to 2.94\%.
This is contextualized by the choice of the triggering objective used for model optimization.
The CSMPO model is trained with a smaller value (15) of the Huber smoothing parameter $\delta$ as compared to the SMPO (25), both optimized independently for maximizing TPR. 
Despite this, the comparable AUC, which competes against TPR in the $\delta$ optimization scan, alludes to the flexible hyperparameter and architecture space of the CSMPO, enabling a balanced approach to optimizing for both AUC and TPR. %~\tcg{Is there some application where we can benefit from both? Dynamic threshold tuning? Online-offline model compatiblity?}
%\tcr{JG: However, its performance in TPR is reduced considerably; comment  about choice of quantization, Huber delta etc that influence this and leave the takeaway that with some engineering one could optimize it for both? Emphasize flexibility!}

To aid in mapping out the benefits of flexibility in the CSMPO architecture for both performance and computational overhead, an alternate CSMPO model is assessed with structure $19\to2\to1$.
For this alternate, the first layer is an SMPO reducing the MPS length from 19 to 2 sites through a spacing of 18, and the second layer reduces it further from 2 to 1. 
A spacing of 18 in the first layer of the CSMPO demonstrates the most severe reduction in free model parameters from the SMPO; the alternate CSMPO has only 264 trainable parameters, a 72\% compression with respect to the SMPO.
This steep compression also lets the model benefit from a symmetric and sparse structure around the middle site allowing for highly parallelizable horizontal contraction implementation. 
The physical and bond dimensions of this model are the same as that of CSMPO$_{19\to7\to1}$.
Table~\ref{tab:result_perf} summarizes the performance metrics for the SMPO and CSMPO models, including both CSMPO options. 
The impact of the extreme compression of the alternate CSMPO can be observed through its reduced performance in both AUC and TPR compared to the primary CSMPO, which will be shown to trade off in greater deployment efficiency in the following section. 

\begin{table}[t]
\centering
\begin{minipage}{\linewidth}
\centering
\begin{tabular}{c|c|ccc}
\multirow{2}{*}{\bf Model} & \multirow{2}{*}{\bf Signal} & \multicolumn{2}{c}{\bf Metric} \\
\hhline{~~--}
 & & AUC [\%] & TPR [\%] \\
\hline
\multirow{4}{*}{SMPO} 
    & $A\rightarrow 4\ell$ & 0.90 & 6.35 \\
    & $h^\pm \rightarrow \tau\nu$ &   0.88   &    0.10  \\
    &  $h^0 \rightarrow \tau\tau$ &   0.80   &    0.11  \\
    & LQ$\rightarrow b\tau$ &  0.86    &   0.10   \\
\hline
\multirow{4}{*}{CSMPO$_{19\to7\to1}$} 
    & $A\rightarrow 4\ell$ & 0.89 & 2.94 \\
    & $h^\pm \rightarrow \tau\nu$ &   0.86   &   0.14   \\
    & $h^0 \rightarrow \tau\tau$ &   0.76   &   0.11   \\
    & LQ$\rightarrow b\tau$ &   0.81   &   0.12   \\
\hline
\multirow{4}{*}{CSMPO$_{19\to2\to1}$} 
    & $A\rightarrow 4\ell$ & 0.78 & 1.23 \\
    & $h^\pm \rightarrow \tau\nu$ &   0.81   &  0.14    \\
    & $h^0 \rightarrow \tau\tau$ &    0.67  &   0.08  \\
    & LQ$\rightarrow b\tau$ &   0.71   &    0.04  \\
\hline
\end{tabular}
\end{minipage}
\caption{Performance for the SMPO and CSMPO models assessed by two metrics, AUC and TPR at FPR = 10$^{-5}$, for the four test signals, considering the best model in the ensemble.}
\label{tab:result_perf}
\end{table}

%------------------------------------------------------
\subsection{FPGA Implementation}

The choice of FPGAs as a deployment target for quantum-inspired ML in edge scenarios is motivated by their widespread availability and their ability to deliver higher efficiency than GPUs.
Furthermore, FPGAs are an integral part of data acquisition systems at the LHC, where they run first-stage trigger algorithms in the $\mathcal{O}$($\mu$s) latencies required to keep up with the LHC bunch crossing rate. 
For example, the ATLAS hardware trigger in Run 3 is comprised of Xilinx Ultrascale+ devices with a total latency of 2.5 $\mu$s, which is expected to increase to  10 $\mu$s  in the upgraded system for the HL-LHC~\cite{Aad_2024}.

\subsubsection{Quantization}
Quantization of model inputs and weights is a common approach to compressing models for hardware deployment in resource-constrained scenarios.
However, the reduction of precision can impact the model performance, introducing a trade-off that can be assessed by the scanning and testing of multiple quantization levels. 
For this study, the standard 32-bit floating point precision was tested against nine other fixed-point values, with the least precise being 12-bit. 

\begin{figure}[!ht]
\centering
\subfloat[$\|\mathbf{MPS}\|^2$\label{fig:smpo_mps}]
{\includegraphics[width=0.48\textwidth]{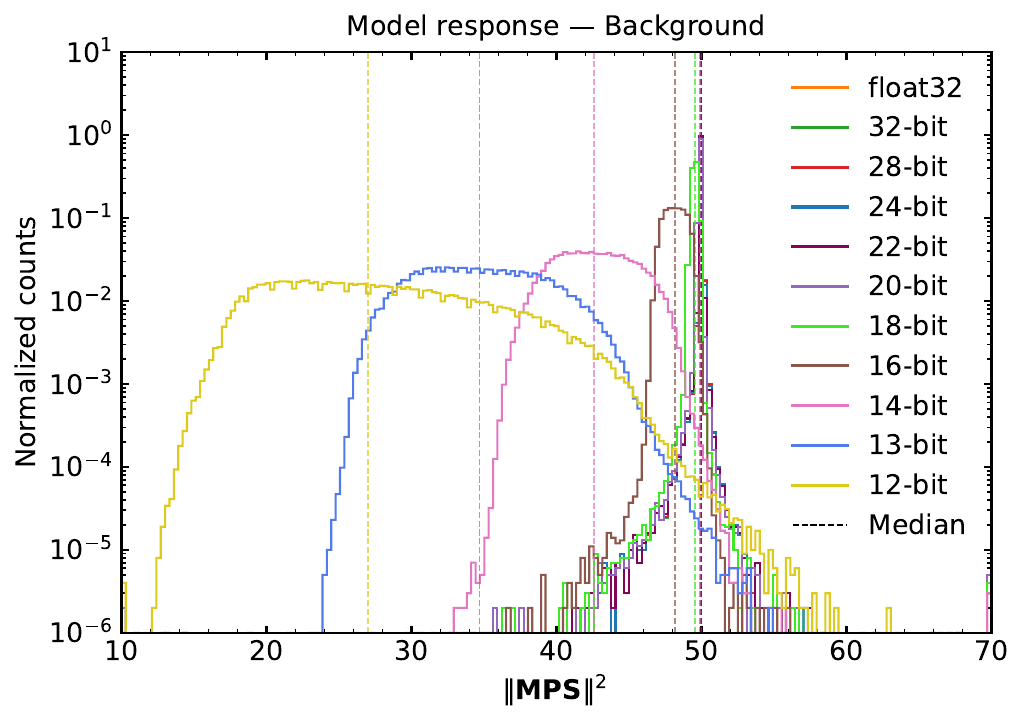}}
\subfloat[Performance vs. quantization \label{fig:smpo_perf_quant}]
{\includegraphics[width=0.48\textwidth]{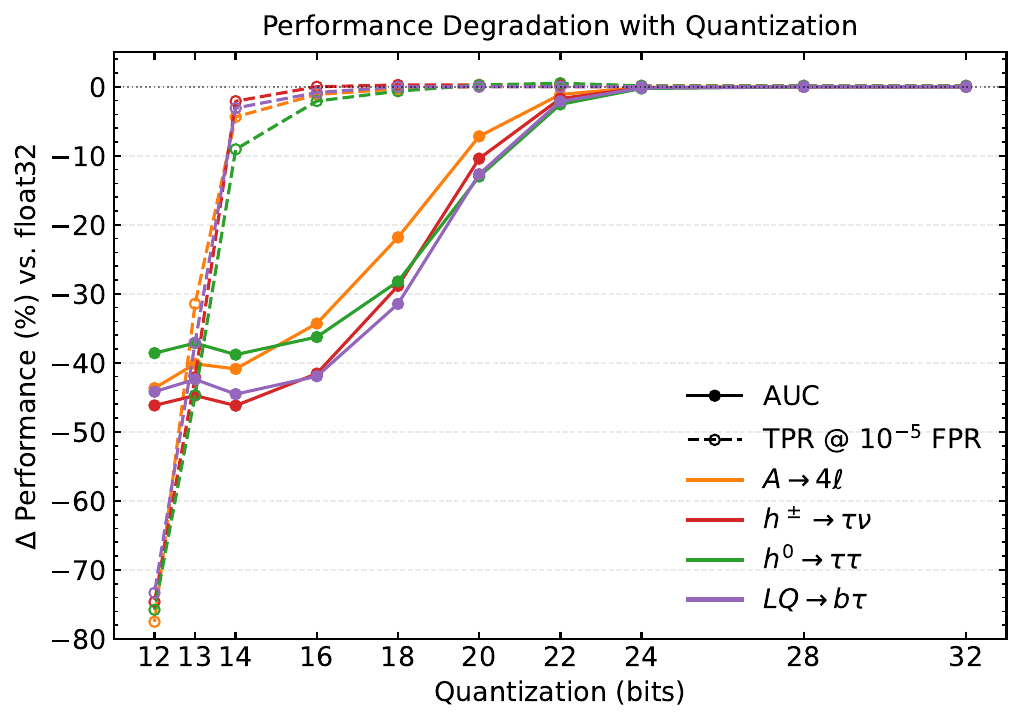}}
\caption{Impact of quantizing model inputs and weights to several fixed-point levels on $||\mathbf{MPS}||^2$ normalized distribution of the background testing events (left), as well as AUC and TPR (right).}
\label{fig:smpo_quant}
\end{figure}

Figure~\ref{fig:smpo_quant} shows the impact of these quantization levels on the anomaly score shape, as well as performance, described by the percent reduction in both AUC and TPR. 
The quantization changes the model response in two important ways.
First, it changes the shape of the anomaly score distribution, namely by broadening the distribution and reducing its long tails. 
This is a relevant factor for the trigger application as it affects the low FPR regime where the model is expected to operate.
Second, quantization shifts the median of the distribution, requiring a recalibration of the target used to define the anomaly score.
The reduction in precision has effectively no impact on performance until the 22-bit level for AUC, and 16-bit level for TPR. 
As this study focuses on the trigger application, 16-bit is chosen for model implementation. 

Most of the mathematical operations in the application of the model are done using the \texttt{ap\_fixed<16,6>} data type. 
The final squared norm calculation uses the data format \texttt{ap\_fixed<16,8,AP\_TRN,AP\_SAT>}, to increase the acceptance range to $[-2^7,2^7)$. 
Saturation clips the maximum norm squared values of events to $2^7$, a threshold found to be sufficiently high to avoid impacting model performance at an FPR of $10^{-5}$. %JG: I don't understand this sentence. SA: I rephrased it. Any better? What I mean to say if an event has such large norms, it doesnt matter if its $2^7$ or any higher. It's large enough to be called anomalous even at FPR of $10^{-5}$.} 

\subsubsection{Tensor Contraction}

The first step in the application of an SMPO on an MPS is the contractions along the physical legs. These contractions are hereon referred to as \textit{vertical} contractions since the physical legs are laid vertically in the tensor diagrams in this work. After the vertical contractions, intermediate SMPO sites with now contracted physical input legs but no physical output legs have no free (or uncontracted) legs. Hence, these sites contract \textit{horizontally} along the bond legs with their neighbor sites until a site with a free leg is encountered. For the various dimensions of the tensor networks used here, the resource critical path is the horizontal contraction. There are various equivalent algorithms to implement these horizontal contractions, which use different levels of FPGA resources. The algorithm choice for this study targets latency minimization. 

\begin{figure}[!ht]
   \centering
   %% Physical index contraction — MPS (top) contracts with SMPO (bottom)
   \subfloat[Vertical Contraction]{%
   \begin{tikzpicture}[scale=.65]
      \tikzset{
         site/.style={circle, inner sep=0pt, outer sep=0pt, minimum width=.4cm},
         osite/.style={site, fill=red!30},
      }
      % MPS row (y=2)
      \foreach \i in {0,...,18} {
         \node[draw, site] (m\i) at (\i, 1.5) {};
      }
      % MPS bonds (trivial, no label)
      \foreach \i in {0,...,17} {
         \pgfmathtruncatemacro{\j}{\i+1}
         \draw[thick] (m\i) -- (m\j);
      }
      % SMPO row (y=0)
      \foreach \i in {0,...,8} {
         \node[draw, site] (s\i) at (\i, 0) {};
      }
      % SMPO row (y=0)
      \foreach \i in {10,...,18} {
         \node[draw, site] (s\i) at (\i, 0) {};
      }
      \node[draw, osite] (s9) at (9, 0) {};
      % SMPO bonds
      \foreach \i in {0,...,17} {
         \pgfmathtruncatemacro{\j}{\i+1}
         \draw[thick] (s\i) -- (s\j);
      }
      % SMPO bond label (first bond)
      \path (s0) -- (s1) node[midway, below=0.15cm] {$b$};
      % Physical index contraction (double-headed arrows)
      \foreach \i in {0,...,18} {
         \draw[<->, thick] (m\i) -- (s\i);
      }
      % p_i label (on first arrow)
      \path (m0) -- (s0) node[midway, left=0.1cm] {$p_i$};
      % SMPO output index
      \draw[thick] (s9) -- ++(0,-1) node[below] {$p_o$};
   \end{tikzpicture}
   \label{fig:contraction-vert}%
   }\\

   %% 19 sites — first horizontal contraction from both ends
   \subfloat[Horizontal Contraction - Step 1]{%
   \begin{tikzpicture}[scale=.65]
      \tikzset{
         site/.style={circle, inner sep=0pt, outer sep=0pt, minimum width=.4cm},
         osite/.style={site, fill=red!30},
         csite/.style={site, fill=green!30},
      }
      % Green endpoints
      \node[draw, csite] (s0) at (0,0) {};
      \node[draw, csite] (s18) at (18,0) {};
      % Regular sites
      \foreach \i in {1,...,8} {
         \node[draw, site] (s\i) at (\i, 0) {};
      }
      \foreach \i in {10,...,17} {
         \node[draw, site] (s\i) at (\i, 0) {};
      }
      % Output site
      \node[draw, osite] (s9) at (9,0) {};
      % Regular bonds
      \foreach \i in {1,...,16} {
         \pgfmathtruncatemacro{\j}{\i+1}
         \draw[thick] (s\i) -- (s\j);
      }
      % Contracting bonds (arrows)
      \draw[<->, thick] (s0) -- (s1);
      \draw[<->, thick] (s17) -- (s18);
      % Output index
      \draw[thick] (s9) -- ++(0,-0.8);
   \end{tikzpicture}
   \label{fig:contraction-hori-1}%
   }

   %% (c) 17 sites — second horizontal contraction
   \subfloat[Horizontal Contraction - Step 2]{%
   \begin{tikzpicture}[scale=.65]
      \tikzset{
         site/.style={circle, inner sep=0pt, outer sep=0pt, minimum width=.4cm},
         osite/.style={site, fill=red!30},
         csite/.style={site, fill=green!30},
      }
      % Green endpoints
      \node[draw, csite] (s0) at (0,0) {};
      \node[draw, csite] (s16) at (16,0) {};
      % Regular sites
      \foreach \i in {1,...,7} {
         \node[draw, site] (s\i) at (\i, 0) {};
      }
      \foreach \i in {9,...,15} {
         \node[draw, site] (s\i) at (\i, 0) {};
      }
      % Output site (center of 17 = index 8)
      \node[draw, osite] (s8) at (8,0) {};
      % Regular bonds
      \foreach \i in {1,...,14} {
         \pgfmathtruncatemacro{\j}{\i+1}
         \draw[thick] (s\i) -- (s\j);
      }
      % Contracting bonds (arrows)
      \draw[<->, thick] (s0) -- (s1);
      \draw[<->, thick] (s15) -- (s16);
      % Output index
      \draw[thick] (s8) -- ++(0,-0.8);
   \end{tikzpicture}
   \label{fig:contraction-hori-2}%
   }\\[0.1em]

   %% Ellipsis for intermediate steps
   $\vdots$\\[1em]

    %% 7 sites
   \subfloat[Horizontal Contraction - Step 7]{%
   \begin{minipage}{.9\linewidth}
   \centering    
   \begin{tikzpicture}[scale=.65]
      \tikzset{
         site/.style={circle, inner sep=0pt, outer sep=0pt, minimum width=.4cm},
         osite/.style={site, fill=red!30},
         csite/.style={site, fill=green!30},
      }
      \node[draw, csite] (s0) at (0,0) {};
      \node[draw, site] (s1) at (1,0) {};
      \node[draw, site] (s2) at (2,0) {};
      \node[draw, osite] (s3) at (3,0) {};
      \node[draw, site] (s4) at (4,0) {};
      \node[draw, site] (s5) at (5,0) {};
      \node[draw, csite] (s6) at (6,0) {};
      % Regular bonds
      \draw[thick] (s1) -- (s2) -- (s3) -- (s4) -- (s5);
      % Contracting bonds (arrows)
      \draw[<->, thick] (s0) -- (s1);
      \draw[<->, thick] (s5) -- (s6);
      % Output index
      \draw[thick] (s3) -- ++(0,-0.8);
   \end{tikzpicture}
   \label{fig:contraction-7}%
   \end{minipage}
   }
   \\
   %% 5 sites
   \subfloat[Horizontal Contraction - Step 8]{%
   \begin{minipage}{0.5\linewidth}
   \centering
   \begin{tikzpicture}[scale=.65]
      \tikzset{
         site/.style={circle, inner sep=0pt, outer sep=0pt, minimum width=.4cm},
         osite/.style={site, fill=red!30},
         csite/.style={site, fill=green!30},
      }
      \node[draw, csite] (s0) at (0,0) {};
      \node[draw, site] (s1) at (1,0) {};
      \node[draw, osite] (s2) at (2,0) {};
      \node[draw, site] (s3) at (3,0) {};
      \node[draw, csite] (s4) at (4,0) {};
      % Regular bonds
      \draw[thick] (s1) -- (s2) -- (s3);
      % Contracting bonds (arrows)
      \draw[<->, thick] (s0) -- (s1);
      \draw[<->, thick] (s3) -- (s4);
      % Output index
      \draw[thick] (s2) -- ++(0,-0.8);
   \end{tikzpicture}
   \label{fig:contraction-hori-8}%
   \end{minipage}
   }
   \hfill
   %% (e) 3 sites
   \subfloat[3-site Contraction]{%
   \begin{minipage}{0.25\linewidth}
   \centering
   \begin{tikzpicture}[scale=.65]
      \tikzset{
         site/.style={circle, inner sep=0pt, outer sep=0pt, minimum width=.4cm},
         osite/.style={site, fill=red!30},
         csite/.style={site, fill=green!30},
      }
      \node[draw, csite] (s0) at (0,0) {};
      \node[draw, osite] (s1) at (1,0) {};
      \node[draw, csite] (s2) at (2,0) {};
      % Both bonds are contracting
      \draw[<->, thick] (s0) -- (s1);
      \draw[<->, thick] (s1) -- (s2);
      % Output index
      \draw[thick] (s1) -- ++(0,-0.8);
   \end{tikzpicture}
   \label{fig:contraction-3-site}%
   \end{minipage}
   }
   \hfill
   %% Norm — single tensor contracting with itself
   \subfloat[$\|\mathbf{MPS}\|^2$]{%
   \begin{minipage}{0.2\linewidth}
   \centering
   \begin{tikzpicture}[scale=.65]
      \tikzset{
         site/.style={circle, inner sep=0pt, outer sep=0pt, minimum width=.4cm},
         nsite/.style={site, fill=brown!50},
      }
      \node[draw, nsite] (s0) at (0,0) {};
      \node[draw, nsite] (s1) at (1.5,0) {};
      \draw[<->, thick] (s0) -- (s1) node[midway, below=0.15cm] {$p_o$};
   \end{tikzpicture}
   \label{fig:contraction-norm}%
   \end{minipage}
   }

   \caption{Tensor contraction steps in the implementation of the SMPO model application on an input event. The first step is the vertical contraction along the physical output legs of the input embedded MPS and the physical input legs of the SMPO (a). The horizontal contraction is implemented as a bi-directional contraction chain (b) - (e) between the extreme sites (shaded in green) and their neighbor. The remaining three sites are contracted together (f) to yield the final vector site, which is contracted with itself (g) to yield the squared norm $\|\mathbf{MPS}\|^2$.}
   \label{fig:smpo-application}
\end{figure}
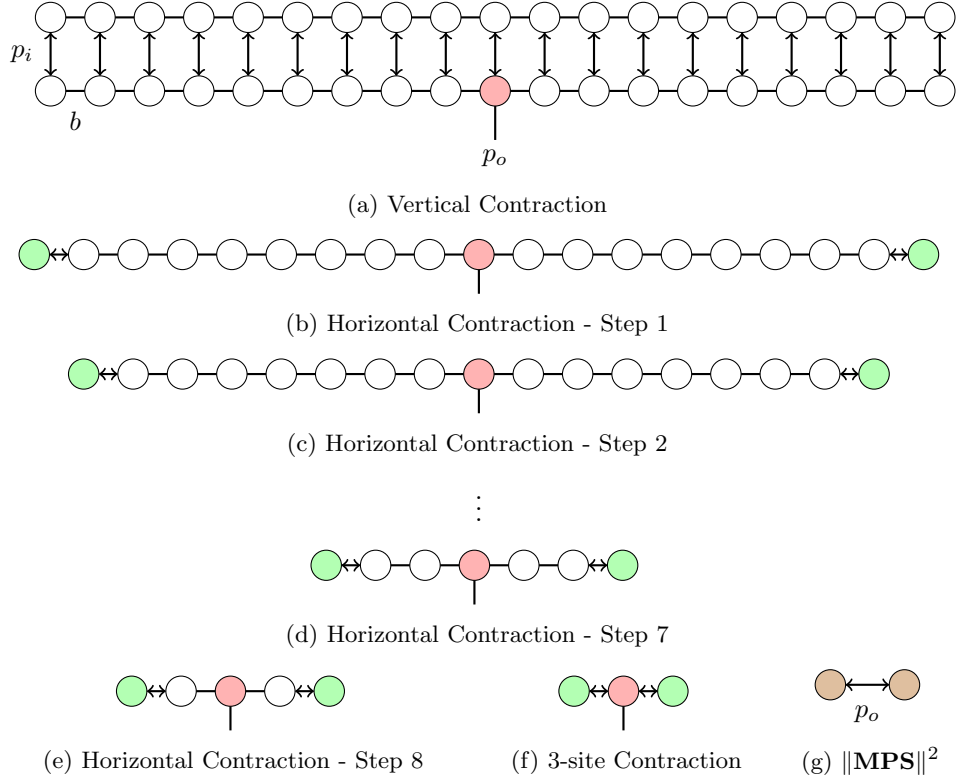

Figure~\ref{fig:smpo-application} shows the tensor contraction steps used in the implementation of the SMPO model application to get the final squared norm used to define the anomaly score. The latency optimal algorithm is chosen to be a sweep from both ends of the tensor network running concurrently, until there are just three tensors left: a left site, a right site, and a middle anchor site which contains the physical output leg. Finally, the three tensors are contracted to yield the trivial MPS, a vector of length three. The vector norm is then used to define the anomaly score.

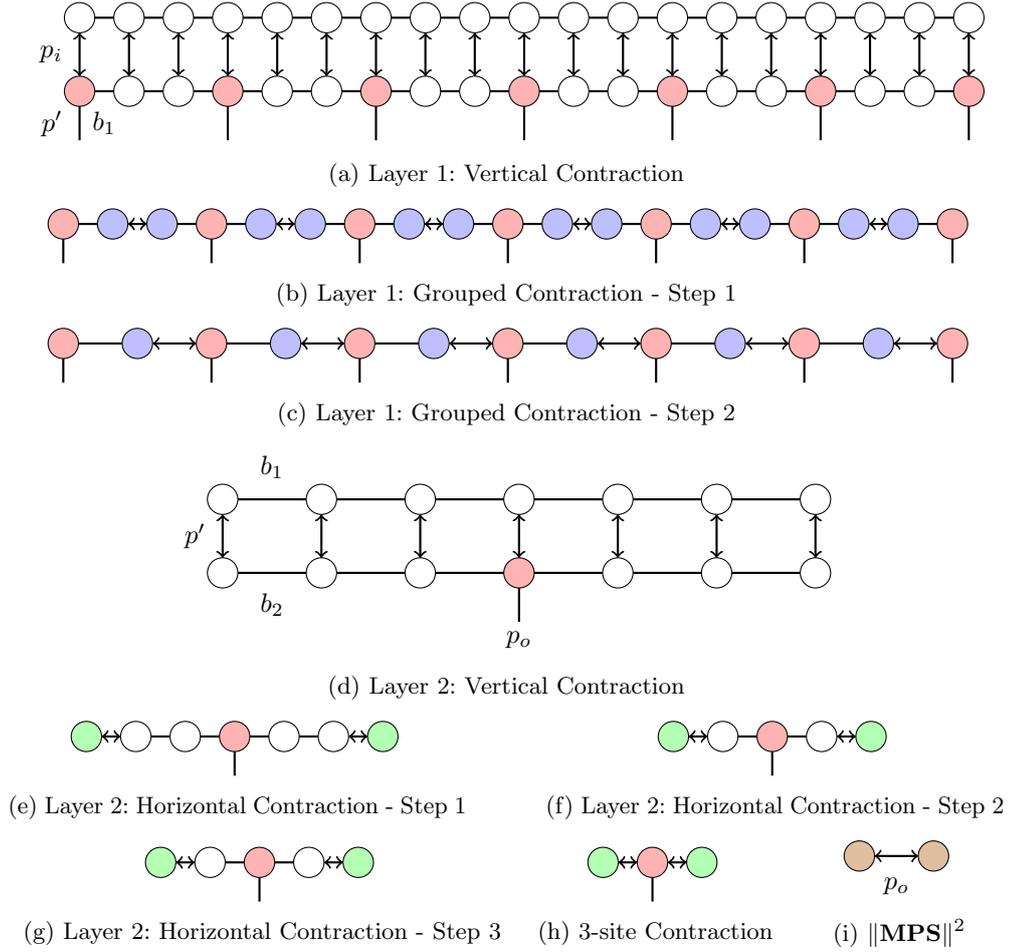
\begin{figure}[!ht]
   \centering

   %% ================================================================
   %% (a) Vertical contraction: MPS (top) with Layer 1 SMPO (bottom)
   %% ================================================================
   \subfloat[Layer 1: Vertical Contraction]{%
   \begin{tikzpicture}[scale=.65]
      \tikzset{
         site/.style={circle, inner sep=0pt, outer sep=0pt, minimum width=.4cm},
         osite/.style={site, fill=red!30},
      }
      % MPS row (y=1.5)
      \foreach \i in {0,...,18} {
         \node[draw, site] (m\i) at (\i, 1.5) {};
      }
      % MPS bonds (trivial, no label)
      \foreach \i in {0,...,17} {
         \pgfmathtruncatemacro{\j}{\i+1}
         \draw[thick] (m\i) -- (m\j);
      }
      % L1 SMPO row (y=0) — non-output sites
      \foreach \i in {1,2,4,5,7,8,10,11,13,14,16,17} {
         \node[draw, site] (s\i) at (\i, 0) {};
      }
      % L1 SMPO row (y=0) — output sites (red)
      \foreach \i in {0,3,6,9,12,15,18} {
         \node[draw, osite] (s\i) at (\i, 0) {};
      }
      % L1 SMPO bonds
      \foreach \i in {0,...,17} {
         \pgfmathtruncatemacro{\j}{\i+1}
         \draw[thick] (s\i) -- (s\j);
      }
      % L1 bond label
      \path (s0) -- (s1) node[midway, below=0.15cm] {$b_1$};
      % Vertical contraction arrows
      \foreach \i in {0,...,18} {
         \draw[<->, thick] (m\i) -- (s\i);
      }
      % p_i label
      \path (m0) -- (s0) node[midway, left=0.1cm] {$p_i$};
      % Output indices from red sites
      \foreach \i in {0,3,6,9,12,15,18} {
         \draw[thick] (s\i) -- ++(0,-1);
      }
      % p' label
      \path (s0) -- ++(0,-1) node[midway, left=0.1cm] {$p'$};
   \end{tikzpicture}
   \label{fig:csmpo-vert-l1}%
   }\\

   %% ================================================================
   %% ( Step 1: Pair contractions within each group
   %% ================================================================
   \subfloat[Layer 1: Grouped Contraction - Step 1]{%
   \begin{tikzpicture}[scale=.65]
      \tikzset{
         site/.style={circle, inner sep=0pt, outer sep=0pt, minimum width=.4cm},
         osite/.style={site, fill=red!30},
         psite/.style={site, fill=blue!25},
      }
      % Output sites (red) at positions 0,3,6,9,12,15,18
      \foreach \i in {0,3,6,9,12,15,18} {
         \node[draw, osite] (s\i) at (\i, 0) {};
      }
      % Non-output sites (blue) — contracted pairs
      \foreach \i in {1,2,4,5,7,8,10,11,13,14,16,17} {
         \node[draw, psite] (s\i) at (\i, 0) {};
      }
      % Regular bonds (between output and adjacent non-output, and output-to-output spans)
      \draw[thick] (s0) -- (s1);
      \draw[thick] (s2) -- (s3);
      \draw[thick] (s3) -- (s4);
      \draw[thick] (s5) -- (s6);
      \draw[thick] (s6) -- (s7);
      \draw[thick] (s8) -- (s9);
      \draw[thick] (s9) -- (s10);
      \draw[thick] (s11) -- (s12);
      \draw[thick] (s12) -- (s13);
      \draw[thick] (s14) -- (s15);
      \draw[thick] (s15) -- (s16);
      \draw[thick] (s17) -- (s18);
      % Contracting pair bonds (arrows)
      \draw[<->, thick] (s1) -- (s2);
      \draw[<->, thick] (s4) -- (s5);
      \draw[<->, thick] (s7) -- (s8);
      \draw[<->, thick] (s10) -- (s11);
      \draw[<->, thick] (s13) -- (s14);
      \draw[<->, thick] (s16) -- (s17);
      % Output indices from red sites
      \foreach \i in {0,3,6,9,12,15,18} {
         \draw[thick] (s\i) -- ++(0,-0.8);
      }
   \end{tikzpicture}
   \label{fig:csmpo-pass1}%
   }\\

   %% ================================================================
   %% ( Step 2: Chain absorption into anchor sites
   %% ================================================================
   \subfloat[Layer 1: Grouped Contraction - Step 2]{%
   \begin{tikzpicture}[scale=.65]
      \tikzset{
         site/.style={circle, inner sep=0pt, outer sep=0pt, minimum width=.4cm},
         osite/.style={site, fill=red!30},
         psite/.style={site, fill=blue!25},
      }
      % Site 0 (red, copied through) — standalone
      \node[draw, osite] (s0) at (0, 0) {};
      \draw[thick] (s0) -- ++(0,-0.8);
      % Group 1: chain(1,2) -> anchor 3
      \node[draw, psite] (c1) at (1.5, 0) {};
      \node[draw, osite] (a1) at (3, 0) {};
      \draw[<->, thick] (c1) -- (a1);
      \draw[thick] (s0) -- (c1);
      \draw[thick] (a1) -- ++(0,-0.8);
      % Group 2: chain(4,5) -> anchor 6
      \node[draw, psite] (c2) at (4.5, 0) {};
      \node[draw, osite] (a2) at (6, 0) {};
      \draw[<->, thick] (c2) -- (a2);
      \draw[thick] (a1) -- (c2);
      \draw[thick] (a2) -- ++(0,-0.8);
      % Group 3: chain(7,8) -> anchor 9
      \node[draw, psite] (c3) at (7.5, 0) {};
      \node[draw, osite] (a3) at (9, 0) {};
      \draw[<->, thick] (c3) -- (a3);
      \draw[thick] (a2) -- (c3);
      \draw[thick] (a3) -- ++(0,-0.8);
      % Group 4: chain(10,11) -> anchor 12
      \node[draw, psite] (c4) at (10.5, 0) {};
      \node[draw, osite] (a4) at (12, 0) {};
      \draw[<->, thick] (c4) -- (a4);
      \draw[thick] (a3) -- (c4);
      \draw[thick] (a4) -- ++(0,-0.8);
      % Group 5: chain(13,14) -> anchor 15
      \node[draw, psite] (c5) at (13.5, 0) {};
      \node[draw, osite] (a5) at (15, 0) {};
      \draw[<->, thick] (c5) -- (a5);
      \draw[thick] (a4) -- (c5);
      \draw[thick] (a5) -- ++(0,-0.8);
      % Group 6: chain(16,17) -> anchor 18
      \node[draw, psite] (c6) at (16.5, 0) {};
      \node[draw, osite] (a6) at (18, 0) {};
      \draw[<->, thick] (c6) -- (a6);
      \draw[thick] (a5) -- (c6);
      \draw[thick] (a6) -- ++(0,-0.8);
   \end{tikzpicture}
   \label{fig:csmpo-pass2}%
   }\\

   %% ================================================================
   %% (d) Vertical contraction: 7-site MPS with Layer 2 SMPO
   %% ================================================================
   \subfloat[Layer 2: Vertical Contraction]{%
   \begin{minipage}{0.9\linewidth}
   \centering
   \begin{tikzpicture}[scale=.65]
      \tikzset{
         site/.style={circle, inner sep=0pt, outer sep=0pt, minimum width=.4cm},
         osite/.style={site, fill=red!30},
      }
      % 7-site MPS row (y=1.5) — result of L1 contraction
      \foreach \i in {0,...,6} {
         \node[draw, site] (m\i) at (\i*2, 1.5) {};
      }
      % MPS bonds
      \foreach \i in {0,...,5} {
         \pgfmathtruncatemacro{\j}{\i+1}
         \draw[thick] (m\i) -- (m\j);
      }
      % L2 SMPO row (y=0) — non-output sites
      \foreach \i in {0,1,2,4,5,6} {
         \node[draw, site] (s\i) at (\i*2, 0) {};
      }
      % L2 SMPO center output site (red)
      \node[draw, osite] (s3) at (6, 0) {};
      % L2 SMPO bonds
      \foreach \i in {0,...,5} {
         \pgfmathtruncatemacro{\j}{\i+1}
         \draw[thick] (s\i) -- (s\j);
      }
      % L2 bond label
      \path (m0) -- (m1) node[midway, above=0.15cm] {$b_1$};
      % L2 bond label
      \path (s0) -- (s1) node[midway, below=0.15cm] {$b_2$};
      % Vertical contraction arrows
      \foreach \i in {0,...,6} {
         \draw[<->, thick] (m\i) -- (s\i);
      }
      % p' label
      \path (m0) -- (s0) node[midway, left=0.1cm] {$p'$};
      % Output index from center
      \draw[thick] (s3) -- ++(0,-1) node[below] {$p_o$};
   \end{tikzpicture}
   \label{fig:csmpo-vert-l2}%
   \end{minipage}
   }\\

   %% ================================================================
   %% (e) 7-site horizontal contraction start
   %% ================================================================
   \subfloat[Layer 2: Horizontal Contraction - Step 1]{%
   \begin{minipage}{0.45\linewidth}
   \centering
   \begin{tikzpicture}[scale=.65]
      \tikzset{
         site/.style={circle, inner sep=0pt, outer sep=0pt, minimum width=.4cm},
         osite/.style={site, fill=red!30},
         csite/.style={site, fill=green!30},
      }
      \node[draw, csite] (s0) at (0,0) {};
      \node[draw, site] (s1) at (1,0) {};
      \node[draw, site] (s2) at (2,0) {};
      \node[draw, osite] (s3) at (3,0) {};
      \node[draw, site] (s4) at (4,0) {};
      \node[draw, site] (s5) at (5,0) {};
      \node[draw, csite] (s6) at (6,0) {};
      % bonds
      \draw[thick] (s1) -- (s2) -- (s3) -- (s4) -- (s5);
      % Contracting bonds with arrows
      \draw[<->, thick,] (s0) -- (s1);
      \draw[<->, thick,] (s5) -- (s6);
      % Output index
      \draw[thick] (s3) -- ++(0,-0.8);
   \end{tikzpicture}
   \label{fig:csmpo-hori-l2-1}%
   \end{minipage}
   }\hfill
   \subfloat[Layer 2: Horizontal Contraction - Step 2]{%
   \begin{minipage}{0.45\linewidth}
   \centering
   \begin{tikzpicture}[scale=.65]
      \tikzset{
         site/.style={circle, inner sep=0pt, outer sep=0pt, minimum width=.4cm},
         osite/.style={site, fill=red!30},
         csite/.style={site, fill=green!30},
      }
      \node[draw, csite] (s0) at (0,0) {};
      \node[draw, site] (s1) at (1,0) {};
      \node[draw, osite] (s2) at (2,0) {};
      \node[draw, site] (s3) at (3,0) {};
      \node[draw, csite] (s4) at (4,0) {};
      % bonds
      \draw[thick] (s1) -- (s2) -- (s3);
      \draw[<->, thick,] (s0) -- (s1);
      \draw[<->, thick,] (s3) -- (s4);
      % Output index
      \draw[thick] (s2) -- ++(0,-0.8);
   \end{tikzpicture}
   \label{fig:csmpo-hori-l2-2}%
   \end{minipage}
   }\\

   %% ================================================================
   %% (f) 5 sites — bonds
   %% ================================================================
   \subfloat[Layer 2: Horizontal Contraction - Step 3]{%
   \begin{minipage}{0.5\linewidth}
   \centering
   \begin{tikzpicture}[scale=.65]
      \tikzset{
         site/.style={circle, inner sep=0pt, outer sep=0pt, minimum width=.4cm},
         osite/.style={site, fill=red!30},
         csite/.style={site, fill=green!30},
      }
      \node[draw, csite] (s0) at (0,0) {};
      \node[draw, site] (s1) at (1,0) {};
      \node[draw, osite] (s2) at (2,0) {};
      \node[draw, site] (s3) at (3,0) {};
      \node[draw, csite] (s4) at (4,0) {};
      % bonds
      \draw[thick] (s1) -- (s2) -- (s3);
      % Contracting bonds
      \draw[<->, thick] (s0) -- (s1);
      \draw[<->, thick] (s3) -- (s4);
      % Output index
      \draw[thick] (s2) -- ++(0,-0.8);
   \end{tikzpicture}
   \label{fig:csmpo-hori-l2-3}%
   \end{minipage}
   }
   \hfill
   %% ================================================================
   %% (g) 3 sites — bonds
   %% ================================================================
   \subfloat[3-site Contraction]{%
   \begin{minipage}{0.25\linewidth}
   \centering
   \begin{tikzpicture}[scale=.65]
      \tikzset{
         site/.style={circle, inner sep=0pt, outer sep=0pt, minimum width=.4cm},
         osite/.style={site, fill=red!30},
         csite/.style={site, fill=green!30},
      }
      \node[draw, csite] (s0) at (0,0) {};
      \node[draw, osite] (s1) at (1,0) {};
      \node[draw, csite] (s2) at (2,0) {};
      % Both bonds contracting
      \draw[<->, thick] (s0) -- (s1);
      \draw[<->, thick] (s1) -- (s2);
      % Output index
      \draw[thick] (s1) -- ++(0,-0.8);
   \end{tikzpicture}
   \label{fig:csmpo-3site}%
   \end{minipage}
   }
   \hfill
   %% ================================================================
   %% (h) Norm — single tensor contracting with itself
   %% ================================================================
   \subfloat[$\|\mathbf{MPS}\|^2$]{%
   \begin{minipage}{0.2\linewidth}
   \centering
   \begin{tikzpicture}[scale=.65]
      \tikzset{
         site/.style={circle, inner sep=0pt, outer sep=0pt, minimum width=.4cm},
         nsite/.style={site, fill=brown!50},
      }
      \node[draw, nsite] (s0) at (0,0) {};
      \node[draw, nsite] (s1) at (1.5,0) {};
      \draw[<->, thick] (s0) -- (s1) node[midway, below=0.15cm] {$p_o$};
   \end{tikzpicture}
   \label{fig:csmpo-norm}%
   \end{minipage}
   }

   \caption{Tensor contraction steps in the implementation of the CSMPO model application on an input event. The first step is the vertical contraction along the physical output legs of the input embedded MPS and the physical input legs of the first layer of the CSMPO (a). Sites without physical output legs (shaded in blue) are contracted with each other (b) followed by contraction to their right neighbor (c). The 7-site MPS created at this stage is vertically contracted with the second layer of the CSMPO (d), followed by a bi-directional contraction chain (e) - (g) between the extreme sites (shaded in green) and their neighbor. The remaining three sites are contracted together (h) to yield the final vector site, which is contracted with itself (i) to yield the squared norm $\|\mathbf{MPS}\|^2$.}
   \label{fig:csmpo-application}
\end{figure}

The CSMPO architecture allows for more flexibility in the horizontal contraction implementation. Figure~\ref{fig:csmpo-application} shows the tensor contraction steps used in the implementation of the CSMPO model application. After the application of the first SMPO layer, the sites with no free legs are contracted with their next neighbors. 
This is done in groups of three, with the first site is left as is, as shown in Figures~\ref{fig:csmpo-pass1} and~\ref{fig:csmpo-pass2}. 
The second layer of the CSMPO is then applied on this intermediate MPS, following which a bi-directional sweep from both ends of the tensor network is used to ultimately yield the vector norm.
The greater variety of evaluation options for the CSMPO offers another highlight of its advantage over SMPOs in a hardware deployment, where the inference can be optimized based on hardware type and application-specific constraints on features such as power or latency. 

\subsubsection{Resource Analysis}

The model application is written in C++ as a high-level synthesis implementation, synthesized to RTL using the AMD Vitis$^\mathrm{TM}$ HLS tool. 
The final SMPO synthesis is simulated to an AMD/Xilinx Kintex UltraScale XCKU115 target FPGA
%~\tcr{part number xcku115 flvb2104 2 i} 
using $5.5\pm0.2$~ns cycle duration.
Table~\ref{tab:results_resources} summarizes the resulting resources for the SMPO, CSMPO, and alternate CSMPO models. 
All three models are capable of FPGA deployment without the use of DSPs, which are often used for MACs of high-precision numbers and thus are typically the most limited resources for ML deployment.
Furthermore, the anomaly detection decision can be made in sub-microsecond latencies, well within the overall latency requirements for current and future collider experiments. 

The CSMPO architecture enables a more compact model, yielding up to a 35\% reduction in latency when implemented on an FPGA.
However, it does lead to a modest increase in LUT usage compared to the SMPO, as the latency-optimised implementation exploits the structural independence of the first layer's sub-contractions by spatially parallelising them on the FPGA, requiring replicated compute units and fully partitioned register storage for intermediate values.
This trade-off again underlines the inherent flexibility of the CSMPO deployment introduced by its distribution of operations across multiple SMPO layers, allowing it to be more easily engineered for specific optimization criteria or data analysis tasks. 
% \tcr{Sagar anything to add here?}
%\tcr{State resources, latency: meets future collider trigger system constraints. 
%JG: Highlighting the AUC and params/MACs makes the CSMPO benefit a more "theoretical/mathematical" result... which is fine as long as we can explain why the FPGA/quantization/Huber delta engineering and TPR worked out the way they did and propose a task-specific engineering solution. Highlight again flexibility offered by CSMPO and ways we can tweak engineering to a given objective}

These results demonstrate the potential of the CSMPO as a new variety of SMPO that maintains strong performance and is particularly conducive to FPGA deployment. 
Compared to SMPOs, the increased number of hyperparameters and greater flexibility in model structure, along with freedom in the choice of contraction algorithm, make CSMPOs more amenable to optimization for specific inference tasks and hardware targets, positioning them as a strong option for deployment in highly tailored HEP data pipelines.
Future efforts could investigate additional ways to maintain performance of compressed CSMPO models, such as the introduction of non-linearity between layers for further enhancement of learning capacity, or further reduce its resource footprint, namely through an analogy of pruning to reduce the number of computations needed for inference. 

\begin{table}[h]
\centering
\begin{minipage}{\linewidth}
\centering
\begin{tabular}{c|ccc}
\hline
Model & SMPO & CSMPO$_{19\rightarrow7\rightarrow1}$ & CSMPO$_{19\rightarrow2\rightarrow1}$ \\ 
\hline 
%TPR & 6.27 & 1.41 & 1.23  \\ 
\textbf{\# Parameters} & \textbf{936} & \textbf{456} & \textbf{264}  \\ 
\textbf{\# MACs} & \textbf{1255} & \textbf{1039} & \textbf{455}  \\ 
LUTs & 76669 & 143754 & 100146 \\ 
FFs & 9817 & 8272 & 4870  \\ 
DSPs & 0 & 0 & 0 \\ 
\textbf{Latency} [$\mu$s] & \textbf{0.37} & \textbf{0.33} & \textbf{0.24}  \\ 
\end{tabular}
\end{minipage}
\caption{Model details, namely number of trainable parameters and number of multiply-and-accumulate (MAC) operations for inference, along with FPGA resources (expressed in LUTs, DSPs, FFs, and latency) comparing the SMPO and CSMPO model benchmarks.}
\label{tab:results_resources}
\end{table}

%%%%%%%%%%%%%%%%%%%%%%%%%%%%%%%%%%%%%%%%%%%%%%%%%%%%%%%%%%%
%%%%%%%%%%%%%%%%%%%%    Conclusions    %%%%%%%%%%%%%%%%%%%%
%%%%%%%%%%%%%%%%%%%%%%%%%%%%%%%%%%%%%%%%%%%%%%%%%%%%%%%%%%%

\section{Conclusions}
\label{sec:conclusions}
Tensor networks, specifically SMPOs, provide a performant and efficient means of learning over high energy particle collision events. 
Furthermore, their inherent linearity and sparsity make them a good candidate for computationally restricted applications such as real-time triggering.
An SMPO model trained over simulated LHC events and optimized for broad signal sensitivity is found to provide anomaly detection performance competitive with state-of-the-art methods.
Custom FPGA implementation of this model achieves resource and latency estimations viable for trigger system designs at current and future colliders.
Furthermore, the cascaded SMPO architecture is found to provide similar learning capacity with enhanced flexibility in model architecture and FPGA deployment, both of which are key features when considering codesign with hardware platforms for implementation at the edge.  
These results realize the use of TNs in today's classical hardware platforms, thereby providing an entry point for the use of quantum-inspired methods in future collider experiments and other real-time scientific applications.

%%%%%%%%%%%%%%%%%%%%%%%%%%%%%%%%%%%%%%%%%%%%%%%%%%%%%%%%%%%
%%%%%%%%%%%%%%%%%%%%    Acknowledgements    %%%%%%%%%%%%%%%
%%%%%%%%%%%%%%%%%%%%%%%%%%%%%%%%%%%%%%%%%%%%%%%%%%%%%%%%%%%

\section*{Acknowledgments}

This work is supported by the U.S. Department of Energy under contract number DE-AC02-76SF00515.

\appendix

\section{Code Availability}
The background and signal samples used in this study are available on Zenodo~\cite{thea_aarrestad_2021_5046389,thea_aarrestad_2021_5046446,thea_aarrestad_2021_5061633,thea_aarrestad_2021_7152617,thea_aarrestad_2021_5055454}, and relevant machine learning code is publicly available at \url{https://github.com/SLAC-Julia-Group/hardware-aware-tensor-networks}.

\section{MAC Calculation}
\label{app:mac}
In the following calculation, a MAC is defined as one multiply-accumulate operation: \texttt{acc += a * b}. Chained products ($a \times b \times c$) are decomposed into their minimum multiply count via intermediate contractions. This ensures fair comparison across architectures regardless of implementation fusion choices.

We adopt the following corner-site conventions, which are critical for accurate parameter counting. 
At the left boundary (site~0), the left bond dimension is fixed to \(1\) and the right bond 
dimension is \(b\); at the right boundary (site~\(N-1\)), the left bond dimension is \(b\) and 
the right bond dimension is \(1\). For non-output sites, the physical output dimension is set 
to \(1\), whereas for output sites the physical output dimension is \(p_o = 3\).

\subsection{Single-layer SMPO}

The architecture of the single-layer SMPO is $19 \to 1$, with bond $b = 4$ and output at site 9. The physical dimension $p_i = 3$ (\texttt{PHYS\_IN}) and $p_o = 3$ (\texttt{PHYS\_OUT}) with the number of input sites $N = 19$. 

There are three steps to SMPO inference that contribute to the total MAC count: vertical contraction, horizontal contraction, and final vector norm.

\textbf{1. Vertical contraction}

Each output element of \texttt{smpo\_out[site][p][l][r]} requires $p_i = 3$ MACs (accumulation over physical input). Element count per site equals the product of its bond dimensions and physical output dimension.

\begin{table}[h!]
\centering
\begin{tabular}{lcc}
\toprule
Site(s) & Elements & MACs \\
\midrule
0       & $1 \times 1 \times 4 = 4$             & 12  \\
1--8    & $8 \times 1 \times 4 \times 4 = 128$  & 384 \\
9       & $3 \times 4 \times 4 = 48$             & 144 \\
10--17  & $8 \times 1 \times 4 \times 4 = 128$  & 384 \\
18      & $1 \times 4 \times 1 = 4$              & 12  \\
\midrule
\textbf{Subtotal} & & \textbf{936} \\
\bottomrule
\end{tabular}
\end{table}

\textbf{2. Horizontal contraction (bidirectional sweep + merge)}

Left sweep --- 8 steps (sites $1 \to 8$). Each step is a vector--matrix multiply:
$\texttt{left\_env}[r] = \sum_b \texttt{left\_env}[b] \cdot T_s[b][r]$. Per step: $b^2 = 16$ MACs.

\begin{table}[h!]
\centering
\begin{tabular}{lcc}
\toprule
Component & Formula & MACs \\
\midrule
Left sweep  & $8 \times b^2$ & 128 \\
Right sweep & $8 \times b^2$ & 128 \\
Merge pass 1: $rc[p][l] = \sum_r T_9[p][l][r] \cdot \texttt{right\_env}[r]$ & $p_o \times b \times b$ & 48 \\
Merge pass 2: $\texttt{output}[p] = \sum_l \texttt{left\_env}[l] \cdot rc[p][l]$ & $p_o \times b$ & 12 \\
\midrule
\textbf{Subtotal} & & \textbf{316} \\
\bottomrule
\end{tabular}
\end{table}

\textbf{3. Norm}: $p_o = \mathbf{3}$ MACs.

Table~\ref{tab:app_mac_smpo} provides a summary of the MACs calculated for each step, indicating a total sum of 1255 for the 19$\rightarrow$1 SMPO. 
\begin{table}[h!]
\centering
\begin{tabular}{lc}
\toprule
Inference step & MACs \\
\midrule
Vertical  & 936  \\
Horizontal  & 316  \\
Norm      & 3    \\
\midrule
\textbf{Total} & \textbf{1255} \\
\bottomrule
\end{tabular}
\caption{MAC count by inference step for the SMPO model. \label{tab:app_mac_smpo}}
\end{table}

%---------------------------------------------------------
%---------------------------------------------------------
\subsection{Cascaded SMPO}
%---------------------------------------------------------
%---------------------------------------------------------

The cascaded architectures follow the format $19 \to M \to 1$ with $b_1 = b_2 = 2$ and composite bond $B = b_1 b_2 = 4$.
The input is a 19-site product state with bond dimension \(1\) and physical dimension \(p_i=3\). 
The intermediate output is an \(M\)-site MPS with bond dimension \(b_1=2\) and physical dimension \(p'=3\).
Output sites are placed symmetrically. For \(M=2\), the output sites are \(\{0,\,18\}\), and for \(M=7\), 
the output sites are \(\{0,\,3,\,6,\,9,\,12,\,15,\,18\}\).

%%%%%%%%%%%%%%%%
\subsubsection{Cascade Layer 1: $19 \to M$ SMPO, $b_1 = 2$}

%-------- 
\textbf{1. Vertical contraction}

As with the SMPO, each output element costs $p_i = 3$ MACs. 

\begin{table}[h!]
\centering
\begin{tabular}{lccc}
\toprule
Site(s) & Role & Elements & MACs \\
\midrule
0      & Left boundary, output   & $3 \times 1 \times 2 = 6$            & 18  \\
1--17  & Interior, non-output    & $17 \times 1 \times 2 \times 2 = 68$ & 204 \\
18     & Right boundary, output  & $3 \times 2 \times 1 = 6$            & 18  \\
\midrule
\textbf{Subtotal} & & & \textbf{240} \\
\bottomrule
\end{tabular}
\caption{$M = 2$, outputs at $\{0,\, 18\}$ \label{app:tab_1vertx2}}
\end{table}

\begin{table}[h!]
\centering
\begin{tabular}{lccc}
\toprule
Site(s) & Role & Elements & MACs \\
\midrule
0      & Left boundary, output   & 6  & 18  \\
1--2   & Interior, non-output    & 8  & 24  \\
3      & Interior, output        & 12 & 36  \\
4--5   & Interior, non-output    & 8  & 24  \\
6      & Interior, output        & 12 & 36  \\
7--8   & Interior, non-output    & 8  & 24  \\
9      & Interior, output        & 12 & 36  \\
10--11 & Interior, non-output    & 8  & 24  \\
12     & Interior, output        & 12 & 36  \\
13--14 & Interior, non-output    & 8  & 24  \\
15     & Interior, output        & 12 & 36  \\
16--17 & Interior, non-output    & 8  & 24  \\
18     & Right boundary, output  & 6  & 18  \\
\midrule
\textbf{Subtotal} & & & \textbf{360} \\
\bottomrule
\end{tabular}
\caption{$M = 7$, outputs at $\{0,\, 3,\, 6,\, 9,\, 12,\, 15,\, 18\}$ \label{app:tab_1vertx7}}
\end{table}

%-------- 
\par\noindent\textbf{2. Horizontal contraction (forming the $M$-site output MPS)}

Between consecutive output sites, $g$ non-output interior sites form a chain of $[b_1] \times [b_1]$ matrices that must be contracted and absorbed into the adjacent output tensor.

\underline{Decomposition (minimum multiply count per group of $g$ non-output sites)}

\textbf{Step 1 --- Chain}: Contract the $g$ non-output matrices into a single $[b_1][b_1]$ matrix via $(g-1)$ sequential matrix--matrix products. Each product costs $b_1^3$ MACs (standard $b_1 \times b_1$ matrix multiply).

\textbf{Step 2 --- Absorb}: Contract the chain result with the adjacent output site tensor. Produces $[p'][b_1][b_1]$ output elements, each requiring a sum over $b_1$ terms:
\begin{equation}
  \texttt{out}[p][l][r] = \sum_{b_1} D[l][b_1] \times C[p][b_1][r], \qquad
  \text{Cost: } p' \times b_1^2 \times b_1 = p' \times b_1^3 \text{ MACs}.
\end{equation}

Per group total: $(g-1) \times b_1^3 + p' \times b_1^3 = b_1^3(g - 1 + p')$.

\vspace{\baselineskip}

\underline{Closed-form (summing over all $M - 1$ groups)}
    
\begin{equation}
  L1_\text{horiz} = b_1^3 \times \bigl[(N - 2M + 1) + (M-1)\,p'\bigr] = 136 + 8M
  \quad (b_1 = 2,\; p' = 3,\; N = 19).
\end{equation}

\begin{table}[h!]
\centering
\begin{tabular}{llcccr}
\toprule
Group & Sites & $g$ & Chain MACs & Absorb MACs & Total \\
\midrule
$0 \to 18$ & 1--17 & 17 & 128 & 24 & 152 \\
\midrule
\textbf{Subtotal} & & & & & \textbf{152} \\
\bottomrule
\end{tabular}
\caption{$M = 2$, 1 group ($g = 17$) \label{app:tab_1horix2}}
\end{table}

\begin{table}[h!]
\centering
\begin{tabular}{llcccr}
\toprule
Group & Sites & $g$ & Chain MACs & Absorb MACs & Total \\
\midrule
$0 \to 3$   & 1, 2   & 2 & 8 & 24 & 32 \\
$3 \to 6$   & 4, 5   & 2 & 8 & 24 & 32 \\
$6 \to 9$   & 7, 8   & 2 & 8 & 24 & 32 \\
$9 \to 12$  & 10, 11 & 2 & 8 & 24 & 32 \\
$12 \to 15$ & 13, 14 & 2 & 8 & 24 & 32 \\
$15 \to 18$ & 16, 17 & 2 & 8 & 24 & 32 \\
\midrule
\textbf{Subtotal} & & & & & \textbf{192} \\
\bottomrule
\end{tabular}
\caption{$M = 7$, 6 groups \label{app:tab_1horix7}}
\end{table}

%%%%%%%%%%%%%%%%
\subsubsection{Cascade Layer 2: $M \to 1$ SMPO, $b_2 = 2$} 

For the second cascade layer, the input is the \(M\)-site MPS produced by Layer~1, with bond dimension \(b_1=2\) and physical dimension \(p'=3\). 
The output is a one-site vector of length \(p_o=3\), with composite bond dimension \(B=b_1 b_2=4\).

\textbf{1. Vertical Contraction (Composite Indexing)}

The contraction of the MPS physical output leg with the SMPO weight produces output tensors with composite bond indices:
\begin{equation}
  \texttt{out}[p_{\text{out},\text{eff}}][l_\text{mps} \cdot l_\text{smpo}][r_\text{mps} \cdot r_\text{smpo}]
  = \sum_{p'} \texttt{MPS}[p'][l_\text{mps}][r_\text{mps}]
    \times W[p'][p_{\text{out},\text{eff}}][l_\text{smpo}][r_\text{smpo}].
\end{equation}

Each output element requires $p_o = 3$ MACs. Composite bond dimensions per site account for both MPS and SMPO boundary conditions:
\begin{itemize}
  \item Left boundary: $B_L = b_{1,L}^\text{(mps)} \times b_{2,L}^\text{(smpo)} = 1 \times 1 = 1$, $\;B_R = b_1 \times b_2 = 4$
  \item Interior: $B_L = B_R = b_1 \times b_2 = 4$
  \item Right boundary: $B_L = b_1 \times b_2 = 4$, $\;B_R = 1 \times 1 = 1$
\end{itemize}

\begin{table}[h!]
\centering
\begin{tabular}{lccccc}
\toprule
Site & $p_{\text{out},\text{eff}}$ & $B_L$ & $B_R$ & Elements & MACs \\
\midrule
0 (output, left boundary)       & 3 & 4 & 1 & 12 & 36 \\
1 (non-output, right boundary)  & 1 & 1 & 4 & 4  & 12 \\
\midrule
\multicolumn{5}{l}{\textbf{Subtotal}} & \textbf{48} \\
\bottomrule
\end{tabular}
\caption{$M = 2$ (output at site 0) \label{app:tab_2vertx2}}
\end{table}

\begin{table}[h!]
\centering
\begin{tabular}{lccccc}
\toprule
Site & $p_{\text{out},\text{eff}}$ & $B_L$ & $B_R$ & Elements & MACs \\
\midrule
0 (non-output, left boundary)  & 1 & 1 & 4 & 4  & 12  \\
1 (non-output, interior)       & 1 & 4 & 4 & 16 & 48  \\
2 (non-output, interior)       & 1 & 4 & 4 & 16 & 48  \\
3 (output, interior)           & 3 & 4 & 4 & 48 & 144 \\
4 (non-output, interior)       & 1 & 4 & 4 & 16 & 48  \\
5 (non-output, interior)       & 1 & 4 & 4 & 16 & 48  \\
6 (non-output, right boundary) & 1 & 4 & 1 & 4  & 12  \\
\midrule
\multicolumn{5}{l}{\textbf{Subtotal}} & \textbf{360} \\
\bottomrule
\end{tabular}
\caption{$M = 7$ (output at site 3) \label{app:tab_2vertx7}}
\end{table}

\textbf{Horizontal Contraction (at Composite Bond $B = 4$)}

Identical structure to the single $19 \to 1$ horizontally, but over $M$ sites instead of 19, at composite bond $B = 4$. Output site at center: $\lfloor M-1/2 \rfloor$.

Left/right wing sizes: $w = \lfloor M-1/2 \rfloor$ sweep steps per wing.

\begin{table}[h!]
\centering
\begin{tabular}{lcc}
\toprule
Component & Formula & MACs \\
\midrule
Left sweep  & 0 steps & 0  \\
Right sweep & 0 steps & 0  \\
Merge: $\texttt{output}[p] = \sum_r T_1[p][0][r] \cdot \texttt{right\_env}[r] $ & $p_o \times B$ & 12 \\
\midrule
\textbf{Subtotal} & & \textbf{12} \\
\bottomrule
\end{tabular}
\caption{$M = 2$ (output at site 1). Left wing: site 0 only (boundary init, 0 sweep steps).
Right wing: none (output is left boundary). \label{app:tab_2horix2}}
\end{table}

\begin{table}[h!]
\centering
\begin{tabular}{lcc}
\toprule
Component & Formula & MACs \\
\midrule
Left sweep     & $2 \times B^2$  & 32 \\
Right sweep    & $2 \times B^2$  & 32 \\
Merge pass 1   & $p_o \times B^2$ & 48 \\
Merge pass 2   & $p_o \times B$   & 12 \\
\midrule
\textbf{Subtotal} & & \textbf{124} \\
\bottomrule
\end{tabular}
\caption{$M = 7$ (output at site 3). Left wing: sites 0 (init), 1, 2 (sweep) $\to$ 2 sweep steps.
Right wing: sites 6 (init), 5, 4 (sweep) $\to$ 2 sweep steps. \label{app:tab_2horix7}}
\end{table}

\subsection{Summary}

\begin{table}[h!]
\centering
\begin{tabular}{lccc}
\toprule
Component & $M = 2$ & $M = 7$ & Single \\
\midrule
L1 Vertical   & 240 & 360 & 936  \\
L1 Horizontal & 152 & 192 & ---  \\
L2 Vertical   & 48  & 360 & ---  \\
L2 Horizontal & 12  & 124 & 316  \\
Norm          & 3   & 3   & 3    \\
\midrule
\textbf{Total} & \textbf{455}  & \textbf{1039} & \textbf{1255} \\
vs.\ single    & $-63.7\%$     & $-17.2\%$     & ---           \\
\bottomrule
\end{tabular}
\end{table}

The cascade's MAC advantage is structural:

\begin{enumerate}
  \item \textbf{L1 Vertical}: 19 sites operate at $b_1^2 = 4$ instead of $b^2 = 16$ (4$\times$ cheaper per site). This alone saves ${\sim}600$ MACs.

  \item \textbf{L2 Vertical}: Only $M$ sites (not 19) operate at composite $B^2 = 16$. Each additional L2 site costs $p' \times B^2 = 48$ MACs --- the dominant cost-driver for large $M$.

  \item \textbf{L1 Horizontal}: Chain contractions scale as $b_1^3 = 8$ per product, plus a fixed $p' \times b_1^3 = 24$ absorb cost per group. The absorb cost makes L1 horizontal monotonically increasing in $M$ (more groups = more absorb overhead), but the total is modest relative to L2 savings.

  \item \textbf{L2 Horizontal}: Fewer sites means fewer sweep steps. $M = 2$ has zero sweep steps; $M = 7$ has only 2 per wing.
\end{enumerate}

The fundamental tradeoff: decreasing $M$ reduces L2 cost (fewer sites at expensive $B^2 = 16$ operations) but increases L1 horizontal cost (longer chains, though at cheap $b_1^3 = 8$ per product). Since $p' \times B^2 = 48$ per additional L2 site vastly exceeds $b_1^3 = 8$ per chain product, reducing $M$ always wins on MACs.

L1 horizontal is increasing in $M$ due to the absorb cost:
\begin{equation}
  \text{total} = b_1^3 \times \bigl[(N - 2M + 1) + (M-1)\,p'\bigr] = 136 + 8M.
\end{equation}
Each new output site adds one absorb at $p' \times b_1^3 = 24$ while removing one chain product at $b_1^3 = 8$, for a net increase of 16 MACs.

%%%%%%%%%%%%%%%%%%%%%%%%%%%%%%%%%%%%%%%%%%%%%%%%%%%%%%%%%%%
%%%%%%%%%%%%%%%%%%%%    Bibliography    %%%%%%%%%%%%%%%%%%%
%%%%%%%%%%%%%%%%%%%%%%%%%%%%%%%%%%%%%%%%%%%%%%%%%%%%%%%%%%%

% \section*{References}

%\bibliographystyle{unsrt}
\bibliography{mybib.bib}

@article{Preskill_2018,
   title={Quantum Computing in the NISQ era and beyond},
   volume={2},
   ISSN={2521-327X},
   url={http://dx.doi.org/10.22331/q-2018-08-06-79},
   DOI={10.22331/q-2018-08-06-79},
   journal={Quantum},
   publisher={Verein zur Forderung des Open Access Publizierens in den Quantenwissenschaften},
   author={Preskill, John},
   year={2018},
   month=aug, pages={79} }

@article{Biamonte_2017,
   title={Quantum machine learning},
   volume={549},
   ISSN={1476-4687},
   url={http://dx.doi.org/10.1038/nature23474},
   DOI={10.1038/nature23474},
   number={7671},
   journal={Nature},
   publisher={Springer Science and Business Media LLC},
   author={Biamonte, Jacob and Wittek, Peter and Pancotti, Nicola and Rebentrost, Patrick and Wiebe, Nathan and Lloyd, Seth},
   year={2017},
   month=sep, pages={195–202} }

@article{doi:10.1126/science.abn7293,
author = {Hsin-Yuan Huang  and Michael Broughton  and Jordan Cotler  and Sitan Chen  and Jerry Li  and Masoud Mohseni  and Hartmut Neven  and Ryan Babbush  and Richard Kueng  and John Preskill  and Jarrod R. McClean },
title = {Quantum advantage in learning from experiments},
journal = {Science},
volume = {376},
number = {6598},
pages = {1182-1186},
year = {2022},
doi = {10.1126/science.abn7293},
URL = {https://www.science.org/doi/abs/10.1126/science.abn7293}
}

@inproceedings{gandrakota2025realtime,
    title        = {{Real-time Anomaly Detection at the L1 Trigger of CMS Experiment}},
    author       = {Gandrakota, A. and {CMS Collaboration}},
    year         = 2025,
    booktitle    = {Proceedings of Science (ICHEP2024)},
    pages        = 1025,
    note         = {PoS ICHEP2024 (2025) 1025}
}

@techreport{Sugizaki:2947542,
      author        = "Sugizaki, Kaito and ATLAS Collaboration",
      collaboration = "ATLAS",
      title         = "{GELATO: A Generic Event-Level Anomalous Trigger Option
                       for ATLAS in LHC Run 3}",
      institution   = "CERN",
      reportNumber  = "ATL-DAQ-PROC-2025-020",
      address       = "Geneva",
      year          = "2025",
      url           = "https://cds.cern.ch/record/2947542",
}

@article{Li2025,
  author       = {Li, Weikang and Ma, Yixuan and Deng, Dong-Ling},
  title        = {Pitfalls and prospects of quantum machine learning},
  journal      = {Nature Computational Science},
  year         = {2025},
  volume       = {5},
  number       = {12},
  pages        = {1095--1097},
  doi          = {10.1038/s43588-025-00914-6},
  url          = {https://doi.org/10.1038/s43588-025-00914-6},
  issn         = {2662-8457}
  }

@article{BELIS2024100091,
title = {Machine learning for anomaly detection in particle physics},
journal = {Reviews in Physics},
volume = {12},
pages = {100091},
year = {2024},
issn = {2405-4283},
doi = {https://doi.org/10.1016/j.revip.2024.100091},
url = {https://www.sciencedirect.com/science/article/pii/S2405428324000017},
author = {Vasilis Belis and Patrick Odagiu and Thea Klaeboe Aarrestad}
}

@article{Deiana_2022,
   title={{Applications and Techniques for Fast Machine Learning in Science}},
   volume={5},
   ISSN={2624-909X},
   url={http://dx.doi.org/10.3389/fdata.2022.787421},
   DOI={10.3389/fdata.2022.787421},
   journal={Frontiers in Big Data},
   publisher={Frontiers Media SA},
   author={Deiana, Allison McCarn and Tran, Nhan and Agar, Joshua and Blott, Michaela and Di Guglielmo, Giuseppe and Duarte, Javier and Harris, Philip and Hauck, Scott and Liu, Mia and Neubauer, Mark S. and Ngadiuba, Jennifer and Ogrenci-Memik, Seda and Pierini, Maurizio and Aarrestad, Thea and Bähr, Steffen and Becker, Jürgen and Berthold, Anne-Sophie and Bonventre, Richard J. and Müller Bravo, Tomás E. and Diefenthaler, Markus and Dong, Zhen and Fritzsche, Nick and Gholami, Amir and Govorkova, Ekaterina and Guo, Dongning and Hazelwood, Kyle J. and Herwig, Christian and Khan, Babar and Kim, Sehoon and Klijnsma, Thomas and Liu, Yaling and Lo, Kin Ho and Nguyen, Tri and Pezzullo, Gianantonio and Rasoulinezhad, Seyedramin and Rivera, Ryan A. and Scholberg, Kate and Selig, Justin and Sen, Sougata and Strukov, Dmitri and Tang, William and Thais, Savannah and Unger, Kai Lukas and Vilalta, Ricardo and von Krosigk, Belina and Wang, Shen and Warburton, Thomas K.},
   year={2022},
   month=apr }

@misc{jiang2024machinelearningevaluationglobal,
      title={{Machine learning evaluation in the Global Event Processor FPGA for the ATLAS trigger upgrade}}, 
      author={Zhixing Jiang and Scott Hauck and Dennis Yin and Bowen Zuo and Ben Carlson and Shih-Chieh Hsu and Allison Deiana and Rohin Narayan and Santosh Parajuli and Jeff Eastlack},
      year={2024},
      eprint={2406.12875},
      archivePrefix={arXiv},
      primaryClass={physics.ins-det},
      url={https://arxiv.org/abs/2406.12875}, 
}

@article{Feynman1982,
  author    = {Richard P. Feynman},
  title     = {Simulating Physics with Computers},
  journal   = {International Journal of Theoretical Physics},
  volume    = {21},
  number    = {6},
  pages     = {467--488},
  year      = {1982},
  doi       = {10.1007/BF02650179},
  url       = {https://doi.org/10.1007/BF02650179},
  issn      = {1572-9575}
}

@book{Aberle:2749422,
      author        = "{{O. Aberle et al}}",
      title         = "{High-Luminosity Large Hadron Collider (HL-LHC): Technical
                       design report}",
      publisher     = "CERN",
      address       = "Geneva",
      series        = "CERN Yellow Reports: Monographs",
      year          = "2020",
      doi           = "10.23731/CYRM-2020-0010",
}

@ARTICLE{10.3389/fphy.2022.888078,
AUTHOR={Benedikt, Michael  and Zimmermann, Frank },
TITLE={{Future Circular Collider: Integrated Programme and Feasibility Study}},
JOURNAL={Frontiers in Physics},
VOLUME={Volume 10 - 2022},
YEAR={2022},
URL={https://www.frontiersin.org/journals/physics/articles/10.3389/fphy.2022.888078},
DOI={10.3389/fphy.2022.888078}, 
ISSN={2296-424X}
}

@article{Guglielmo_2021,
   title={{A Reconfigurable Neural Network ASIC for Detector Front-End Data Compression at the HL-LHC}},
   volume={68},
   ISSN={1558-1578},
   url={http://dx.doi.org/10.1109/TNS.2021.3087100},
   DOI={10.1109/tns.2021.3087100},
   number={8},
   journal={IEEE Transactions on Nuclear Science},
   publisher={Institute of Electrical and Electronics Engineers (IEEE)},
   author={Guglielmo, Giuseppe Di and Fahim, Farah and Herwig, Christian and Valentin, Manuel Blanco and Duarte, Javier and Gingu, Cristian and Harris, Philip and Hirschauer, James and Kwok, Martin and Loncar, Vladimir and Luo, Yingyi and Miranda, Llovizna and Ngadiuba, Jennifer and Noonan, Daniel and Ogrenci-Memik, Seda and Pierini, Maurizio and Summers, Sioni and Tran, Nhan},
   year={2021},
   month=aug, pages={2179–2186} }

@misc{gonski2026machinelearningheterogeneousedge,
      title={{Machine Learning on Heterogeneous, Edge, and Quantum Hardware for Particle Physics (ML-HEQUPP)}}, 
      author={{Julia Gonski et al}},
      year={2026},
      eprint={2602.22248},
      archivePrefix={arXiv},
      primaryClass={physics.ins-det},
      url={https://arxiv.org/abs/2602.22248}, 
      note={Submitted to Phys. Rev. X Intelligence}
}

@article{Duffy2025,
  author    = {Duffy, Callum and Hassanshahi, Mohammad and Jastrzebski, Marcin and Malik, Sarah},
  title     = {Unsupervised beyond-standard-model event discovery at the LHC with a novel quantum autoencoder},
  journal   = {Quantum Machine Intelligence},
  year      = {2025},
  volume    = {7},
  number    = {1},
  pages     = {41},
  month     = mar,
  doi       = {10.1007/s42484-025-00258-4},
  url       = {https://doi.org/10.1007/s42484-025-00258-4},
  issn      = {2524-4914}
}

@article{Huggins_2019,
doi = {10.1088/2058-9565/aaea94},
url = {https://doi.org/10.1088/2058-9565/aaea94},
year = {2019},
month = {jan},
publisher = {IOP Publishing},
volume = {4},
number = {2},
pages = {024001},
author = {Huggins, William and Patil, Piyush and Mitchell, Bradley and Whaley, K Birgitta and Stoudenmire, E Miles},
title = {Towards quantum machine learning with tensor networks},
journal = {Quantum Science and Technology}
}

@article{Ospanov:2022fke,
    author = "Ospanov, Rustem and Feng, Changqing and Dong, Wenhao and Feng, Wenhao and Zhang, Kan and Yang, Shining",
    title = "{{Development of a resource-efficient FPGA-based neural network regression model for the ATLAS muon trigger upgrades}}",
    eprint = "2201.06288",
    archivePrefix = "arXiv",
    primaryClass = "physics.ins-det",
    doi = "10.1140/epjc/s10052-022-10521-8",
    journal = "Eur. Phys. J. C",
    volume = "82",
    number = "6",
    pages = "576",
    year = "2022"
}

@misc{apresyan2023detectorrdneedsgeneration,
      title={{Detector R\&D needs for the next generation $e^+e^-$ collider}}, 
      author={A. Apresyan and M. Artuso and J. Brau and H. Chen and M. Demarteau and Z. Demiragli and S. Eno and J. Gonski and P. Grannis and H. Gray and O. Gutsche and C. Haber and M. Hohlmann and J. Hirschauer and G. Iakovidis and K. Jakobs and A. J. Lankford and C. Pena and S. Rajagopalan and J. Strube and C. Tully and C. Vernieri and A. White and G. W. Wilson and S. Xie and Z. Ye and J. Zhang and B. Zhou},
      year={2023},
      eprint={2306.13567},
      archivePrefix={arXiv},
      primaryClass={hep-ex},
      url={https://arxiv.org/abs/2306.13567}, 
}

@misc{doe_brn,
  title={Basic Research Needs for High Energy Physics Detector Research \& Development},
  year = {2019},
  month = {dec},
  url = {https://science.osti.gov/-/media/hep/pdf/Reports/2020/DOE_Basic_Research_Needs_Study_on_High_Energy_Physics.pdf}
}

@article{PhysRevLett.126.062001,
  title = {Quantum Algorithm for High Energy Physics Simulations},
  author = {Nachman, Benjamin and Provasoli, Davide and de Jong, Wibe A. and Bauer, Christian W.},
  journal = {Phys. Rev. Lett.},
  volume = {126},
  issue = {6},
  pages = {062001},
  numpages = {6},
  year = {2021},
  month = {Feb},
  publisher = {American Physical Society},
  doi = {10.1103/PhysRevLett.126.062001},
  url = {https://link.aps.org/doi/10.1103/PhysRevLett.126.062001}
}

@article{Toledo-Marin2025,
  author       = {Toledo-Mar{\'i}n, J. Quetzalc{\'o}atl and Gonzalez, Sebastian and Jia, Hao and Lu, Ian and Sogutlu, Deniz and Abhishek, Abhishek and Gay, Colin and Paquet, Eric and Melko, Roger G. and Fox, Geoffrey C. and Swiatlowski, Maximilian and Fedorko, Wojciech},
  title        = {Conditioned quantum-assisted deep generative surrogate for particle-calorimeter interactions},
  journal      = {npj Quantum Information},
  year         = {2025},
  volume       = {11},
  number       = {1},
  pages        = {114},
  doi          = {10.1038/s41534-025-01040-x},
  url          = {https://doi.org/10.1038/s41534-025-01040-x},
  issn         = {2056-6387}
  }

@article{ORUS2014117,
title = {A practical introduction to tensor networks: Matrix product states and projected entangled pair states},
journal = {Annals of Physics},
volume = {349},
pages = {117-158},
year = {2014},
issn = {0003-4916},
doi = {https://doi.org/10.1016/j.aop.2014.06.013},
url = {https://www.sciencedirect.com/science/article/pii/S0003491614001596},
author = {Román Orús},
}

@article{tns_sim,
author = {Huang, Cupjin and Zhang, Fang and Newman, Michael and Ni, Xiaotong and Ding, Dawei and Cai, Junjie and Gao, Xun and Wang, Tenghui and Wu, Feng and Zhang, Gengyan and Ku, Hsiang-Sheng and Tian, Zhengxiong and Wu, Junyin and Xu, Haihong and Yu, Huanjun and Yuan, Bo and Szegedy, Mario and Shi, Yaoyun and Zhao, Hui Hai and Chen, Jianxin},
year = {2021},
month = {09},
pages = {1-10},
title = {Efficient parallelization of tensor network contraction for simulating quantum computation},
volume = {1},
journal = {Nature Computational Science},
doi = {10.1038/s43588-021-00119-7}
}

@Article{10.21468/SciPostPhysCore.4.1.001,
	title={{On the descriptive power of Neural-Networks as constrained Tensor Networks with exponentially large bond dimension}},
	author={Mario Collura and Luca Dell'Anna and Timo Felser and Simone Montangero},
	journal={SciPost Phys. Core},
	volume={4},
	pages={001},
	year={2021},
	publisher={SciPost},
	doi={10.21468/SciPostPhysCore.4.1.001},
	url={https://scipost.org/10.21468/SciPostPhysCore.4.1.001},
}

@article{tnsforml,
author = {Shi-Ju Ran  and Gang Su },
title = {Tensor Networks for Interpretable and Efficient Quantum-Inspired Machine Learning},
journal = {Intelligent Computing},
volume = {2},
number = {},
pages = {0061},
year = {2023},
doi = {10.34133/icomputing.0061},
URL = {https://spj.science.org/doi/abs/10.34133/icomputing.0061}
}

@article{Verstraete01032008,
author = {F. Verstraete and V. Murg and J.I. Cirac},
title = {Matrix product states, projected entangled pair states, and variational renormalization group methods for quantum spin systems},
journal = {Advances in Physics},
volume = {57},
number = {2},
pages = {143--224},
year = {2008},
publisher = {Taylor \& Francis},
doi = {10.1080/14789940801912366},
URL = {https://doi.org/10.1080/14789940801912366},
eprint = { https://doi.org/10.1080/14789940801912366}
}

@article{Felser:2020mka,
    author = "Felser, Timo and Trenti, Marco and Sestini, Lorenzo and Gianelle, Alessio and Zuliani, Davide and Lucchesi, Donatella and Montangero, Simone",
    title = "{Quantum-inspired machine learning on high-energy physics data}",
    eprint = "2004.13747",
    archivePrefix = "arXiv",
    primaryClass = "stat.ML",
    doi = "10.1038/s41534-021-00443-w",
    journal = "npj Quantum Inf.",
    volume = "7",
    pages = "111",
    year = "2021"
}

@article{Montangero:2021puw,
    author = "Montangero, Simone and Rico, Enrique and Silvi, Pietro",
    title = "{Loop-free tensor networks for high-energy physics}",
    eprint = "2109.11842",
    archivePrefix = "arXiv",
    primaryClass = "quant-ph",
    doi = "10.1098/rsta.2021.0065",
    journal = "Phil. Trans. A. Math. Phys. Eng. Sci.",
    volume = "380",
    number = "2216",
    pages = "20210065",
    year = "2021"
}

@inproceedings{Acharya2022QubitSeriation,
  author       = {Acharya, Atithi and Rudolph, Manuel and Chen, Jing and Miller, Jacob and Perdomo-Ortiz, Alejandro},
  title        = {Qubit seriation: Improving data-model alignment using spectral ordering},
  booktitle    = {Proceedings of the Machine Learning and the Physical Sciences Workshop at NeurIPS 2022},
  year         = {2022},
  month        = nov,
  note         = {Workshop at the 36th Conference on Neural Information Processing Systems (NeurIPS 2022)},
  eprint       = {2211.15978},
  archivePrefix= {arXiv},
  primaryClass = {quant-ph}
}

@article{Araz:2021zwu,
    author = "Araz, Jack Y. and Spannowsky, Michael",
    title = "{Quantum-inspired event reconstruction with Tensor Networks: Matrix Product States}",
    eprint = "2106.08334",
    archivePrefix = "arXiv",
    primaryClass = "hep-ph",
    reportNumber = "IPPP/20/114",
    doi = "10.1007/JHEP08(2021)112",
    journal = "JHEP",
    volume = "08",
    pages = "112",
    year = "2021"
}

@inproceedings{NIPS2016_5314b967,
 author = {Stoudenmire, Edwin and Schwab, David J},
 booktitle = {Advances in Neural Information Processing Systems},
 editor = {D. Lee and M. Sugiyama and U. Luxburg and I. Guyon and R. Garnett},
 pages = {},
 publisher = {Curran Associates, Inc.},
 title = {{Supervised Learning with Tensor Networks}},
 volume = {29},
 year = {2016},
address = {New York USA}
}

@article{Liu_2023,
   title={Tensor networks for unsupervised machine learning},
   volume={107},
   ISSN={2470-0053},
   url={http://dx.doi.org/10.1103/PhysRevE.107.L012103},
   DOI={10.1103/physreve.107.l012103},
   number={1},
   journal={Physical Review E},
   publisher={American Physical Society (APS)},
   author={Liu, Jing and Li, Sujie and Zhang, Jiang and Zhang, Pan},
   year={2023},
   month=jan }

@article{Wang:2020dvu,
    author = "Wang, Jinhui and Roberts, Chase and Vidal, Guifre and Leichenauer, Stefan",
    title = "{Anomaly Detection with Tensor Networks}",
    eprint = "2006.02516",
    archivePrefix = "arXiv",
    primaryClass = "cs.LG",
    month = "6",
    year = "2020"
}

@misc{puljak2025tn4mltensornetworktraining,
      title={tn4ml: Tensor Network Training and Customization for Machine Learning}, 
      author={Ema Puljak and Sergio Sanchez-Ramirez and Sergi Masot-Llima and Jofre Vallès-Muns and Artur Garcia-Saez and Maurizio Pierini},
      year={2025},
      eprint={2502.13090},
      archivePrefix={arXiv},
      primaryClass={cs.LG},
      url={https://arxiv.org/abs/2502.13090}, 
}

@misc{borella2024ultralowlatencyquantuminspiredmachine,
      title={{Ultra-low latency quantum-inspired machine learning predictors implemented on FPGA}}, 
      author={Lorenzo Borella and Alberto Coppi and Jacopo Pazzini and Andrea Stanco and Marco Trenti and Andrea Triossi and Marco Zanetti},
      year={2024},
      eprint={2409.16075},
      archivePrefix={arXiv},
      primaryClass={hep-ex},
      url={https://arxiv.org/abs/2409.16075}, 
}

@article{puljak2025tensornetworkanomalydetection,
doi = {10.1088/2632-2153/ae0243},
url = {https://doi.org/10.1088/2632-2153/ae0243},
year = {2025},
month = {oct},
publisher = {IOP Publishing},
volume = {6},
number = {4},
pages = {045001},
author = {Puljak, Ema and Pierini, Maurizio and Garcia-Saez, Artur},
title = {{Tensor network for anomaly detection in the latent space of proton collision events at the LHC}},
journal = {Machine Learning: Science and Technology}
}

@misc{coppi2026tensornetworkmodelslowlatency,
      title={{Towards Tensor Network Models for Low-Latency Jet Tagging on FPGAs}}, 
      author={Alberto Coppi and Ema Puljak and Lorenzo Borella and Daniel Jaschke and Enrique Rico and Maurizio Pierini and Jacopo Pazzini and Andrea Triossi and Simone Montangero},
      year={2026},
      eprint={2601.10801},
      archivePrefix={arXiv},
      primaryClass={cs.LG},
      url={https://arxiv.org/abs/2601.10801}, 
}

@article{Govorkova_2022,
   title={{Autoencoders on field-programmable gate arrays for real-time, unsupervised new physics detection at 40 MHz at the Large Hadron Collider}},
   volume={4},
   ISSN={2522-5839},
   url={http://dx.doi.org/10.1038/s42256-022-00441-3},
   DOI={10.1038/s42256-022-00441-3},
   number={2},
   journal={Nature Machine Intelligence},
   publisher={Springer Science and Business Media LLC},
   author={Govorkova, Ekaterina and Puljak, Ema and Aarrestad, Thea and James, Thomas and Loncar, Vladimir and Pierini, Maurizio and Pol, Adrian Alan and Ghielmetti, Nicolò and Graczyk, Maksymilian and Summers, Sioni and Ngadiuba, Jennifer and Nguyen, Thong Q. and Duarte, Javier and Wu, Zhenbin},
   year={2022},
   month=feb, pages={154–161} }

@article{govorkova2021lhcphysicsdatasetunsupervised,
  author    = {Ekaterina Govorkova and Ema Puljak and Thea Aarrestad and Maurizio Pierini and Kinga Anna Woźniak and Jennifer Ngadiuba},
  title     = {{LHC physics dataset for unsupervised New Physics detection at 40 MHz}},
  journal   = {Scientific Data},
  year      = {2022},
  volume    = {9},
  number    = {1},
  pages     = {118},
  doi       = {10.1038/s41597-022-01187-8},
  url       = {https://doi.org/10.1038/s41597-022-01187-8},
  issn      = {2052-4463},
  month     = {Mar}
}

@article{RevModPhys.82.277,
  title = {Colloquium: Area laws for the entanglement entropy},
  author = {Eisert, J. and Cramer, M. and Plenio, M. B.},
  journal = {Rev. Mod. Phys.},
  volume = {82},
  issue = {1},
  pages = {277--306},
  numpages = {0},
  year = {2010},
  month = {Feb},
  publisher = {American Physical Society},
  doi = {10.1103/RevModPhys.82.277},
  url = {https://link.aps.org/doi/10.1103/RevModPhys.82.277}
}

@misc{kingma2017adammethodstochasticoptimization,
      title={Adam: A Method for Stochastic Optimization}, 
      author={Diederik P. Kingma and Jimmy Ba},
      year={2017},
      eprint={1412.6980},
      archivePrefix={arXiv},
      primaryClass={cs.LG},
      url={https://arxiv.org/abs/1412.6980}, 
}

@article{PhysRevLett.79.5194,
  title = {Negative Entropy and Information in Quantum Mechanics},
  author = {Cerf, N. J. and Adami, C.},
  journal = {Phys. Rev. Lett.},
  volume = {79},
  issue = {26},
  pages = {5194--5197},
  numpages = {0},
  year = {1997},
  month = {Dec},
  publisher = {American Physical Society},
  doi = {10.1103/PhysRevLett.79.5194},
  url = {https://link.aps.org/doi/10.1103/PhysRevLett.79.5194}
}

@book{Nielsen_Chuang_2010, 
   address={Cambridge UK},
   title={Quantum Computation and Quantum Information: 10th Anniversary Edition},
   publisher={Cambridge University Press},
   author={Nielsen, Michael A. and Chuang, Isaac L.},
   year={2010}
}

@article{huber1964robust,
  title={Robust estimation of a location parameter},
  author={Huber, Peter J},
  journal={The Annals of Mathematical Statistics},
  volume={35},
  number={1},
  pages={73--101},
  year={1964},
  publisher={Institute of Mathematical Statistics},
  doi={10.1214/aoms/1177703732}
}

@article{Aad_2024,
doi = {10.1088/1748-0221/19/05/P05063},
url = {https://doi.org/10.1088/1748-0221/19/05/P05063},
year = {2024},
month = {may},
publisher = {IOP Publishing},
volume = {19},
number = {05},
pages = {P05063},
author = {{ATLAS Collaboration}},
title = {{The ATLAS experiment at the CERN Large Hadron Collider: a description of the detector configuration for Run 3}},
journal = {Journal of Instrumentation}
}

@article{PhysRevB.48.10345,
  title = {Density-matrix algorithms for quantum renormalization groups},
  author = {White, Steven R.},
  journal = {Phys. Rev. B},
  volume = {48},
  issue = {14},
  pages = {10345--10356},
  numpages = {0},
  year = {1993},
  month = {Oct},
  publisher = {American Physical Society},
  doi = {10.1103/PhysRevB.48.10345},
  url = {https://link.aps.org/doi/10.1103/PhysRevB.48.10345}
}

@article{PhysRevLett.75.3537,
  title = {Thermodynamic Limit of Density Matrix Renormalization},
  author = {\"Ostlund, Stellan and Rommer, Stefan},
  journal = {Phys. Rev. Lett.},
  volume = {75},
  issue = {19},
  pages = {3537--3540},
  numpages = {0},
  year = {1995},
  month = {Nov},
  publisher = {American Physical Society},
  doi = {10.1103/PhysRevLett.75.3537},
  url = {https://link.aps.org/doi/10.1103/PhysRevLett.75.3537}
}

@misc{thea_aarrestad_2021_5046389,
  author       = {Thea Aarrestad and
                  Ekaterina Govorkova and
                  Jennifer Ngadiuba and
                  Ema Puljak and
                  Maurizio Pierini and
                  Kinga Anna Wozniak},
  title        = {Unsupervised New Physics detection at 40 MHz:
                   Training Dataset
                  },
  month        = jun,
  year         = 2021,
  publisher    = {Zenodo},
  doi          = {10.5281/zenodo.5046389},
}

@misc{thea_aarrestad_2021_5046446,
  author       = {Thea Aarrestad and
                  Ekaterina Govorkova and
                  Jennifer Ngadiuba and
                  Ema Puljak and
                  Maurizio Pierini and
                  Kinga Anna Wozniak},
  title        = {Unsupervised New Physics detection at 40 MHz: $A\to4\ell$ Signal Benchmark Dataset
                  },
  month        = jun,
  year         = 2021,
  publisher    = {Zenodo},
  doi          = {10.5281/zenodo.5046446},
}

@misc{thea_aarrestad_2021_5061633,
  author       = {Thea Aarrestad and
                  Ekaterina Govorkova and
                  Jennifer Ngadiuba and
                  Ema Puljak and
                  Maurizio Pierini and
                  Kinga Anna Wozniak},
  title        = {Unsupervised New Physics detection at 40 MHz: $h^0\to\tau\tau$ Signal Benchmark Dataset
                  },
  month        = jul,
  year         = 2021,
  publisher    = {Zenodo},
  doi          = {10.5281/zenodo.5061633},
}

@misc{thea_aarrestad_2021_7152617,
  author       = {Thea Aarrestad and
                  Ekaterina Govorkova and
                  Jennifer Ngadiuba and
                  Ema Puljak and
                  Maurizio Pierini and
                  Kinga Anna Wozniak},
  title        = {Unsupervised New Physics detection at 40 MHz: $h^+\to\tau\nu$ Signal Benchmark Dataset
                  },
  month        = jul,
  year         = 2021,
  publisher    = {Zenodo},
  version      = {v2},
  doi          = {10.5281/zenodo.7152617},
}

@misc{thea_aarrestad_2021_5055454,
  author       = {Thea Aarrestad and
                  Ekaterina Govorkova and
                  Jennifer Ngadiuba and
                  Ema Puljak and
                  Maurizio Pierini and
                  Kinga Anna Wozniak},
  title        = {Unsupervised New Physics detection at 40 MHz: $LQ\to b\tau$ Signal Benchmark Dataset
                  },
  month        = jul,
  year         = 2021,
  publisher    = {Zenodo},
  doi          = {10.5281/zenodo.5055454},
}

\end{document}